\documentclass[]{fairmeta}

\usepackage{amsmath,amsfonts,bm}

\def\eqref#1{equation~\ref{#1}}

\def\1{\bm{1}}

\def\vs{{\bm{s}}}

\DeclareMathAlphabet{\mathsfit}{\encodingdefault}{\sfdefault}{m}{sl}
\SetMathAlphabet{\mathsfit}{bold}{\encodingdefault}{\sfdefault}{bx}{n}

\newcommand{\R}{\mathbb{R}}

\usepackage{cleveref}
\usepackage{graphicx}
\usepackage{booktabs}
\usepackage{algorithm2e}
\usepackage{diagbox}

\usepackage{multirow}
\usepackage{sidecap}
\usepackage{listings}
\usepackage{subcaption}
\usepackage{wrapfig}

\usepackage[OT2,T1]{fontenc}
\DeclareSymbolFont{cyrletters}{OT2}{wncyr}{m}{n}
\DeclareMathSymbol{\Sha}{\mathalpha}{cyrletters}{"58}

\SetKwComment{Comment}{/* }{ */}
\RestyleAlgo{ruled}

\usepackage{pifont}
\newcommand{\cmark}{\ding{51}}%
\newcommand{\xmark}{\ding{55}}%
\usepackage{enumitem}

\usepackage{rotating}
\usepackage{xspace}  
\usepackage{array}  
\newcolumntype{H}{>{\setbox0=\hbox\bgroup}c<{\egroup}@{}}
\usepackage{tabularx}
\newcolumntype{Y}{>{\RaggedRight\arraybackslash}X}

\renewcommand{\lstlistingname}{Pseudocode}
\def\mypar#1{\noindent\textbf{#1.}\hspace{0mm}}

\makeatletter 
\DeclareRobustCommand\onedot{\futurelet\@let@token\@onedot}
\def\@onedot{\ifx\@let@token.\else.\null\fi\xspace}
\makeatother 
\def\eg{\emph{e.g}\onedot} 
\def\ie{\emph{i.e}\onedot} 
\def\vs{\emph{vs}\onedot} 
\def\wrt{\emph{w.r.t}\onedot}

\newcommand{\imnet}{ImageNet-1k\xspace}
\newcommand{\cc}{CC12M\xspace}
\newcommand{\shst}{S320M\xspace}
\newcommand{\lpl}{LPL\xspace}

\title{Boosting Latent Diffusion with Perceptual Objectives}

\author[1,2]{Tariq Berrada Ifriqi}
\author[1]{Pietro Astolfi}
\author[1]{Melissa Hall}
\author[1]{Marton Havasi}
\author[1]{Yohann Benchetrit}
\author[1,3,4,5]{Adriana Romero-Soriano}
\author[2]{Karteek Alahari}
\author[1]{Michal Drozdzal}
\author[1]{Jakob Verbeek}

\affiliation[1]{FAIR at Meta}
\affiliation[2]{Univ. Grenoble Alpes, Inria, CNRS, Grenoble INP, LJK, France}
\affiliation[3]{McGill University}
\affiliation[4]{Mila, Quebec AI institute}
\affiliation[5]{Canada CIFAR AI chair}

\abstract{Latent diffusion models (LDMs) power state-of-the-art high-resolution generative image models. LDMs learn the data distribution in the latent space of an autoencoder (AE) and produce images by mapping the generated latents into RGB image space using the AE decoder. While this approach allows for efficient model training and sampling, it induces a disconnect between the training of the diffusion model and the decoder, resulting in a loss of detail in the generated images. To remediate this disconnect, we propose to leverage the internal features of the decoder to define a {\it latent perceptual loss} (LPL). This loss encourages the models to create sharper and more realistic images. Our loss can be seamlessly integrated with common autoencoders used in latent diffusion models, and can be applied to different generative modeling paradigms such as DDPM with epsilon and velocity prediction, as well as flow matching. Extensive experiments with models trained on three datasets at 256 and 512 resolution show improved quantitative -- with boosts between 6\% and 20\% in  FID -- and qualitative results when using our perceptual loss.\looseness-1
}

\date{\today}
\correspondence{Tariq Berrada (\email{tariqberrada@meta.com})}

\begin{document}

\maketitle

\begin{figure*}[h]
\def\myim#1#2{\includegraphics[width=39.5mm,height=39.5mm]{figures/CC12M_512/#1/#2.png}}
    \tiny
    \centering
    \setlength\tabcolsep{1pt}
    \renewcommand{\arraystretch}{0.2}
    \begin{tabularx}{\linewidth}{cYYYYYY}
     \begin{sideways} \small{\it  w/o \lpl} \end{sideways}   
     & \myim{no_lpl}{000203}
     & \myim{no_lpl}{000342}  
     & \myim{no_lpl}{000344}
     & \myim{no_lpl}{000108}  
     \\
     \\
     \begin{sideways} \small{\it with \lpl} \end{sideways}   
     & \myim{lpl}{000203}
     & \myim{lpl}{000342}  
     & \myim{lpl}{000344}
     & \myim{lpl}{000108}  
\\
     & {\it  \small A kitten with its paws up.} 
     & {\it  \small A stone wall leads up to a solitary tree at sunset.} 
     & {\it \small Statue of a beautiful maiden hidden among the pretty pink flowers.} 
     & {\it \small Peach and cream brides bouquet for a rustic vintage wedding.} 
     \\
    \end{tabularx}
    \caption{
    {\bf Samples from models trained with and without our latent perceptual loss on CC12M.}
    Samples from our model with latent perceptual loss (bottom) have more detail and realistic textures.
    }
    \label{fig:lpl_cc12m}
\end{figure*}

\section{Introduction}

Latent diffusion models (LDMs)~\citep{ldm} have enabled considerable advances in image generation, and elevated the problem of generative image modeling to a level where it has become available as a technology to the public.
A critical part to this success is to define the generative model in the latent space of an autoencoder (AE), which reduces the  resolution  of the representation over which the model is defined, thereby making it possible to scale diffusion methods to larger datasets, resolutions, and architectures than original pixel-based diffusion models~\citep{dhariwal2021diffusion,sohl-ickstein15icml}.

To train an LDM, all images are first projected into a latent space with the encoder of a pre-trained autoencoder, and then, the diffusion model is optimized directly in the latent space.
Note that when learning the diffusion model the AE decoder is not used -- the diffusion model does not receive any training feedback that would ensure that all latent values reachable by the diffusion process decode to a high quality image. 
This training procedure leads to a disconnect between the diffusion model and the AE decoder, prompting the LDM to produce low quality images that oftentimes lack high frequency image components. Moreover, we note that the latent spaces of pre-trained LDM's autoencoders tend to be highly irregular, in the sense that small changes in the latent space can lead to large changes in the generated images, further exacerbating the autoencoder-diffusion disconnect problem.\looseness-1 

In this work, we propose to alleviate this autoencoder-diffusion disconnect  and propose to include the AE decoder in the training objective of LDM. 
In particular, we introduce latent perceptual loss
(LPL) that acts on the decoder's intermediate features to enrich the training signal of LDM.  
This is similar to the use of perceptual losses for image-to-image translation tasks~\citep{johnson16eccv,zhang2018perceptual}, but we apply this idea in the context of generative modeling and use the feature space of the pre-trained AE decoder rather than that of an external pre-trained discriminative network.

Our latent perceptual loss results in sharper and more realistic images, and leads to better structural consistency than the baseline -- see Figure~\ref{fig:lpl_cc12m}. 
We validate  LPL on three datasets of different sizes -- the commonly used datasets \imnet (1M data points) and \cc (12M data points), 
and additionally  a private dataset \shst (320M data points) -- as well as three generative models formulation -- DDPM~\citep{ho2020denoising} with velocity and epsilon prediction, and conditional flow matching model~\citep{lipman2023flow}. 
In our experiments, we report standard image generative model metrics -- such as FID~\citep{heusel17nips}, CLIPScore~\citep{hessel21emnlp}, as well as Precision and Recall~\citep{precision_recall_distributions,kynkaanniemi19nips}. Our experiments show that the use of LPL leads to consistent performance boosts between 6\% and 20\% in terms of FID. 
Our qualitative analysis further highlights the benefits of LPL, showing images that are sharp and contain high-frequency image details. 
In summary, our contributions are:
\begin{itemize}[noitemsep,topsep=-\parskip,leftmargin=*]
    \item We identify a slight disconnect between latent and pixel-space diffusion which can lead to suboptimal results when training latent diffusion and flow models.
    \item We propose the {\it latent perceptual loss (LPL)}, a  perceptual loss variant leveraging the intermediate feature representation of the autoencoder's decoder.
    \item We present extensive experimental results on the \imnet, \cc, and \shst datasets, demonstrating the benefits of \lpl in boosting the model's quality by 6\% to 20\% in terms of FID.
    \item We show that LPL is effective for a variety of generative model formulations including DDPM and conditional flow matching approaches. 
\end{itemize}      

\section{Related work}

\mypar{Diffusion models} 
The generative modeling landscape has been significantly impacted by diffusion models, surpassing previous state-of-the-art  GAN-based methods~\citep{biggan, stylegan, Karras2019stylegan2, stylegan3}.
Diffusion models offer advantages such as more stable training and better scalability, and were successfully applied to a wide range of applications, including 
image generation~\citep{chen2023pixartalpha,ho2020denoising}, 
video generation~\citep{ho2022video,singer2022makeavideo}, 
music generation~\citep{levy2023controllable,roman23audio}, and 
text generation~\citep{wu2023ardiffusion}.
Various improvements of the framework have been proposed, including different schedulers~\citep{lin2024common, hang2024improvednoiseschedulediffusion}, loss weights~\citep{choi2022perceptionprioritizedtrainingdiffusion, min_snr},  and more recently generalizations of the  framework with flow matching~\citep{lipman2023flow}.
In our work we evaluate the use of our latent perceptual loss in three different training paradigms: DDPM under noise and velocity prediction, as well as flow-based training with the optimal transport path.

\mypar{Latent diffusion} 
Due to the iterative nature of the reverse diffusion process, 
training and sampling diffusion models is computationally demanding, in particular at  high resolution.   
Different approaches have been explored to generate high-resolution content.
For example, \cite{ho22cascaded} used a cascaded approach to progressively add  high-resolution details, by conditioning on previously generated lower resolution images.
A more widely adopted approach is to define the generative model in the latent space induced by a pretrained autoencoder~\citep{ldm}, as previously explored for discrete autoregressive generative models~\citep{esser2020taming}. 
Different architectures have been explored to implement diffusion models in the latent space, including convolutional UNet-based architectures~\citep{ldm,podell2024sdxl}, and more recently transformer-based ones~\citep{Peebles2022DiT,chen2023pixartalpha,gao2023masked,sd3} which show better scaling performance.
Working in a lower-resolution latent space accelerates training and inference, but training models using a loss defined in the latent space also deprives them from matching  high-frequency details from the training data distribution.
Earlier approaches to address this problem include the use of a refiner model~\citep{podell2024sdxl}, which consists of a second diffusion model trained on high-resolution high-quality data that is used to noise and denoise the initial latents, similar to how SDEdit works for image editing~\citep{meng2022sdeditguidedimagesynthesis}.  
Our latent perceptual loss addresses this issue in an orthogonal manner by introducing a loss defined across different layers of the AE decoder in the latter stages of the training process. 
Our approach avoids the necessity of training on specialized curated data~\citep{dai23arxiv}, and does not increase the computational cost of inference.

\mypar{Perceptual losses}
The use of internal features of a fixed, pre-trained deep neural network to compare images or image distributions has become common practice as they have been found to correlate  to some extent with human judgement of similarity~\citep{johnson16eccv,zhang2018perceptual}. 
 An example of this is the widely used Fr\'echet Inception Distance (FID) to assess generative image models~\citep{heusel17nips}.
Such ``perceptual'' distances have also been found to be effective as a loss to train networks for image-to-image tasks and  boost image quality as compared to using simple $\ell_1$ or $\ell_2$ reconstruction losses.
They have been used to train  autoencoders~\citep{esser2020taming}, models for  semantic image synthesis~\citep{pix2pix2017,  berrada2024unlockingpretrainedimagebackbones} and super-resolution~\citep{suvorov2021resolution, invest_sr}, and to assess the sample diversity of generative image models~\citep{schonfeld2021you,astolfi2024paretofronts}.
In addition, recent works propose variants that do not require pretrained image backbones~\citep{under_lpips, watson_lpips,veeramacheneni23fwd}. 
 \citet{an2024bring, song2023consistency} employed LPIPS as metric function in pixel space to train cascaded diffusion models and consistency models, respectively.  
 More closely related to our work, \citet{kang2024diffusion2gan} used a perceptual loss ({\it E-LatentLPIPS}) defined in latent space to distill LDMs to conditional GANs, but used a separate image classification network trained over latents rather than the autoencoder's decoder to obtain the features for this loss.
 \cite{lin2024diffusionmodelperceptualloss} added a perceptual loss ({\it Self-Perceptual}) to train LDMs that is defined over the features of the denoiser network in the case of UNet architectures. However, they found this loss to be detrimental when using classifier-free guidance for inference.
In summary, compared to prior work on perceptual losses, our work is different  in that  (i) \lpl is defined over the features of the decoder -- that maps from latent space to RGB pixel space, rather than using a  network that takes RGB images as input -- and (ii) our LPL can be regardlessly applied to train both latent diffusion and flow models and has no architecture constraint in the denoiser model.
\section{Using the latent decoder to define a perceptual loss}
In this section, we analyze the impact of the decoder-diffusion disconnect on the LDM training, and then, we follow with the definition of our latent perceptual loss.

\subsection{Latent diffusion and the $\ell_2$ objective}
We use  $F_\beta$ to refer to an autoencoder that consists of two modules. 
The encoder, $F^e_\beta$, maps RGB images ${\bf x}_0 \in \R^{H \times W \times 3}$ to a latent representation ${\bf z}_0 \in \R^{H/d \times W/d \times C}$, where $d$ is the spatial downscaling factor, and $C$ the channel dimension of the autoencoder.
The decoder, $F^d_\beta$,  maps from the latent space to the RGB image space.
In LDM, the diffusion model, $D_\Theta$, with parameters $\Theta$ is defined over the latent representation of the autoencoder.
We follow a  typical setting, see \eg  \citep{Peebles2022DiT, chen2023pixartalpha,ldm}, where we use a fixed pre-trained autoencoder with a downsampling factor of $d=8$ and a channel capacity of $C=4$.

\mypar{Training Objective}
The diffusion formulation results in an objective  function that is a lower-bound on the log-likelihood of the data.
In the DDPM paradigm~\citep{ho2020denoising}, the variational lower bound can be expressed as a sum of denoising score matching losses~\citep{connection_denoising_auto}, and the objective function can be written as 
$\mathcal{L} = \sum_t \mathcal{L}_t$, where $\mathcal{L}_t = \mathcal{D}_\textrm{KL} \left[ q\left({\bf x}_{t-1} | {\bf x}_t, {\bf x}_0\right) || p_\theta \left({\bf x}_{t-1}|{\bf x}_t\right) \right] = \mathbb{E}_{{\bf x}_0, \pmb{\epsilon}, t} \left[ \frac{\beta_t}{{(1 - \beta_t) (1 - \alpha_t)}} \cdot \lVert { \pmb{\epsilon}} - \epsilon_\theta ({\bf x}_t, t) \rVert^2 \right]$.
\cite{ho2020denoising} observed that disregarding the time step specific weighting resulted in improved sample quality, and introduced a simplified noise reconstruction objective, known as epsilon prediction,
where the objective is the average of the MSE loss between the predicted noise and the noise vector added to the image, 
$\mathcal{L}_\text{simple} = \sum_t \mathbb{E}_{{\bf x}_0, \pmb{\epsilon}, t} \left[ \lVert \pmb{\epsilon} - \epsilon_\theta({\bf x}_t, t) \rVert^2 \right]$. 
The underlying idea is that the better the noise estimation, the better the final sample quality. An equivalent way to interpret this objective is through reparameterization of the target to the original latents,
$
    \mathcal{L}_\text{simple} = \sum_t \lambda_t \cdot \mathbb{E}_{{\bf x}_0, \pmb{\epsilon}, t} \left[ \lVert {\bf x}_0 - \hat{x}_0({\bf x}_t, t; \theta) \rVert^2 \right]
$, where $\hat{x}_0({\bf x}_t, t; \theta) = ({\bf x}_t - \sigma_t \hat{\pmb{\epsilon}}_t)/\alpha_t$ and $\lambda_t = 1/\sigma_t^2$.

We note that the presence of $\ell_2$ in the LDM objective has two 
important implications. First, the $\ell_2$ norm treats all pixels in the latents as equally important and disregards the downstream structure induced by the decoder whose objective is to reconstruct the image from its latents. This is problematic because the autoencoder's latent space has a non-uniform structure and is not equally influenced by the different pixels in the latent code.
Thus, optimizing the $\ell_2$ distance in the diffusion model latent space could be different from optimizing the perceptual distance between images. 
Second, while an $\ell_2$ objective is theoretically justified in the original DDPM formulation, 
generative models trained with an $\ell_2$ reconstruction objective have been observed  to produce blurry images, 
as is the case \eg for VAE models~\citep{kingma2022autoencoding}. 
Such an effect can lead to exposure bias in diffusion (\ie denoiser input mismatch between training and sampling), which has been studied through the lens of the sampling algorithm but not the training objective~\cite{ning2024elucidating}.
The problem of blurry images due to $\ell_2$ reconstruction losses  has been addressed through the use of perceptual losses such as LPIPS~\citep{zhang2018perceptual}, which provide a significant boost to the image quality in settings such as autoencoding~\citep{esser2020taming}, super-resolution~\cite{ledig17cvpr} and image-to-image generative models~\citep{pix2pix2017, park2019SPADE}.\\

To address these two implications of the latent $\ell_2$ optimization, we design a new loss, namely Latent Percetual Loss (LPL), that directly tackles the mismatch between the structures of the latent and the pixel space by matching the target and the predicted latents when partially decoded with the AE decoder. Compared to existing perceptual losses, e.g., LPIPS, that usually rely on features from pre-trained classifier networks, the use of AE decoder features is not standard. However, for LDMs training, the AE decoder provides a perceptually meaningful signal as the intermediate features of the decoder, being closer to pixel space, have a structure more similar to the pixel space. Moreover, these features have higher resolution than the diffusion latent features, which helps in modeling high-level details that reduce blurriness of the predicted image. 

\subsection{Latent perceptual loss}

We propose a loss function that operates  on the features at different depths in the autoencoder's decoder, $F^d_\beta$.
Let ${\bf z}_t = \alpha_t {\bf z}_0 + \sigma_t \pmb{\epsilon}_t$ be a noisy sample in the diffusion model latent space at time $t$, 
and
$\hat{{\bf z}}_0 = ({\bf z}_t - \sigma_t D({\bf z}_t, t; \Theta) )/\alpha_t$  the corresponding estimated noise-free latent at time $t=0$. \\

To enhance the accuracy of the reconstruction process in image space, we propose augmenting the diffusion objective with a penalty term that encourages the predicted forward process distribution to match the reconstructions $\hat{{\bf z}}_0$ at a lower level of compression. This is achieved by introducing an image-level projection penalty on the forward process, which can be expressed as:
\begin{equation} \label{eq:penalty}
    \mathcal{L}_{t-1}^\text{pen} = \mathbb{E}_q \left[ D_\text{KL} \left( q({{ \bf x}}_{t-1}|{{ \bf x}}_t, {{ \bf x}}_0) \parallel p_\Theta({\hat{ \bf x}}_{t-1} | {\hat{ \bf x}}_t) \right) \right].
\end{equation}
In \Cref{app:lpl_rel}, we show that such a penalty, under certain conditions, can be approximated by computing a pairwise distance between the outputs of the projection decoder, which maps latents into samples in image-space.
Guided by this result, our loss term should resemble a reconstruction term between the original image and the predicted/denoised image obtained from the decoder of the autoencoder.
However, this does not address the problem of blurriness exposed in the previous section, which is why 
we opt for a loss structure that is similar to LPIPS, where the loss term is defined over the intermediate feature spaces of the decoder.

\begin{figure}
    \centering
    \includegraphics[width=\linewidth]{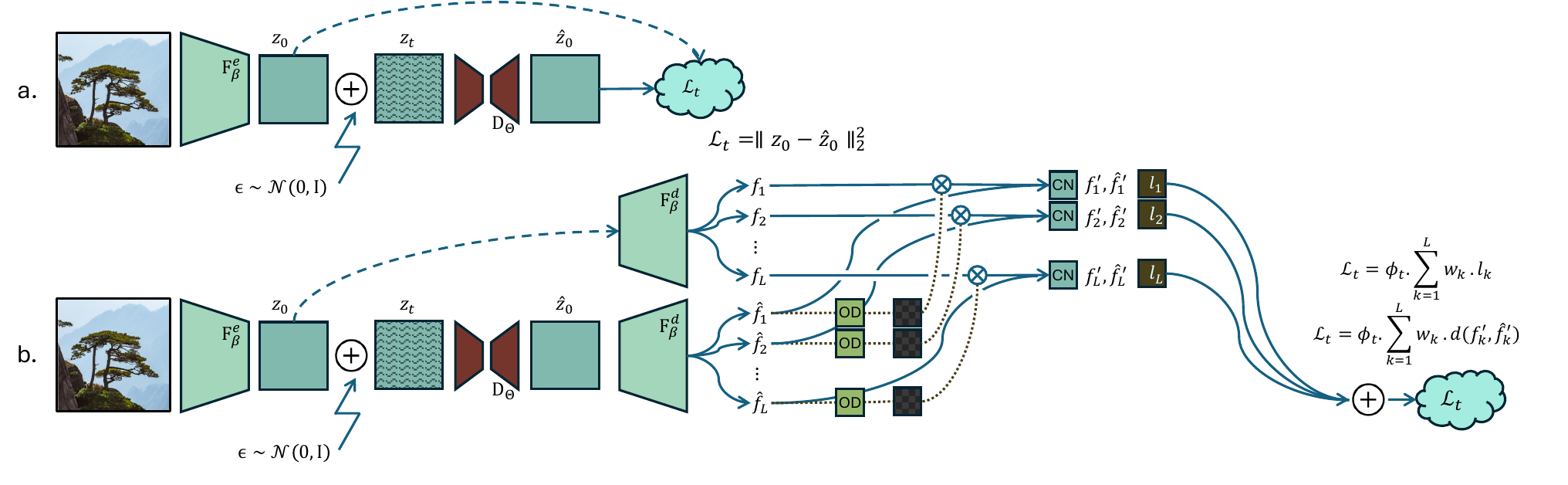}
    \vspace{-2mm}
    \caption{{\bf Overview of our approach.} 
    (a) Latent diffusion models compare clean latents and the predicted latents.
    (b) Our \lpl acts in the features of the autoencoder's decoder effectively aligning the diffusion process with the decoder. 
    {\it $F_\beta^e,F_\beta^d$: autoencoder encoder and decoder, $D_\Theta$: denoiser network, CN: cross normalization layer, OD: outlier detection.}
    }
    \label{fig:main_diagram}
\end{figure}
We compute two sets of hierarchies of $L$ decoder features, $\{\pmb{\phi}_l\}_{l=1}^{L}$, and, $\{\hat{\pmb{\phi}}_l\}_{l=1}^{L}$, by decoding both the original, ${\bf z}_0$, and estimated latents, $\hat{{\bf z}}_0$ (where for brevity we drop the dependence of $\hat{{\bf z}}_0$ on $t$):
\begin{equation}
    \begin{cases}
        \pmb{\phi}_1, ..., \pmb{\phi}_L = \big ( F_\beta^{d, l}({\bf z}_0) \big )_{l \in [\![1, L]\!]},\\
        \hat{\pmb{\phi}}_1, ..., \hat{\pmb{\phi}}_L = \big( F_\beta^{d, l}(\hat{{\bf z}}_0) \big)_{l \in [\![1, L]\!]}.
    \end{cases}
\end{equation}
Using these intermediate features, we can define our training objective. Our LPL, $\mathcal{L}_\textit{\lpl}$, is a weighted sum of the quadratic distances between the feature representations at the different decoding scales, obtained after normalization:

\begin{equation}
\label{eq:lpl_equation1}
    \mathcal{L}_\textrm{LPL} = \mathbb{E}_{t\in \mathcal{T}, \pmb{\epsilon} \sim \mathcal{N}(0,I), {\bf x}_0 \in D_\mathcal{X}} \left[\delta_{\sigma_t \leq \tau_\sigma} \sum_{l=1}^L \frac{\omega_l}{ C_l}  \sum_{c=1}^{C_l} \big\lVert \rho_{l,c}(\hat{\phi}_{l,c}) \odot \big( \pmb{\phi}'_{l, c} - \hat{\pmb{\phi}}'_{l,c} \big) \big\rVert_2^2 \right].
\end{equation}

where $\pmb{\phi}'_l$ is the standardized version of $\pmb{\phi}_l$ across the channel dimension, $\rho_{l,c}(\hat{\pmb{\phi}}_{l,c})$ is a binary map masking the detected outliers in the feature map $\hat{\pmb{\phi}}_{l,c}$, $\omega_l$ is a depth-specific weighting and $C_l$ the channel dimensionality of the feature tensor. Note that we better explain these terms later in this section. Moreover, to reduce both LPL computational complexity and memory overhead, we only apply our loss for high signal-to-noise ratios (SNR). 
In particular, we impose a hard threshold $\tau_\sigma$ and only apply the loss if the SNR is higher than it $\delta_{\sigma_t \leq \tau_\sigma}(\sigma_t)$.
The \lpl loss is applied in conjunction with the standard diffusion loss, resulting in the training objective 

$
    \mathcal{L}_\textrm{tot} = \mathcal{L}_\textrm{Diff} + w_\textrm{LPL} \cdot \mathcal{L}_\textrm{LPL}
$.

\mypar{Depth-specific weighting} Empirically, we find the loss amplitude at different decoder layers to differ significantly -- it grows with a factor of two when considering layers with a factor two increase resolution. To balance the contributions from different decoder layers, we therefore weight them  by the inverse of the upscaling factor \wrt the first  layer, \ie $\omega_l = 2^{-r_l/{r_1}}$ where $r_l$ is the resolution of the $l$-th layer.\looseness-1

\mypar{Normalization}
Since the features in the decoder can have significantly varying statistics from each other, we follow \citet{zhang2018perceptual} and normalize them per channel so that the features in every channel in every layer are zero mean and have  unit variance.
However, normalizing the feature maps corresponding to the original and denoised latents with different statistics can induce nonzero gradients even when the absolute value has been correctly predicted.
To obtain a coherent normalization, we  use the feature statistics from the denoised latents to normalize both tensors.

\begin{SCtable}[50]
    \centering
    {\scriptsize
    \begin{tabular}{lccc}
        \toprule
        {\bf Framework} & $\alpha_t$ & $\sigma_t$ & $\hat{{\bf x}}_0$ \\
        \midrule
        DDPM-$\epsilon_t$ & $\sqrt{\bar{\alpha}_t}$ & $\sqrt{1 - \bar{\alpha}_t}$ & $\big({\bf x}_t - \sigma_t D({\bf x}_t, t; \Theta)\big)/ \alpha_t$\\
        DDPM-$v_\theta$ & $\sqrt{\bar{\alpha}_t}$ & $\sqrt{1 - \bar{\alpha}_t}$ & $\alpha_t {\bf x}_t - \sigma_t D({\bf x}_t, t; \Theta)$\\
        \midrule
        Flow-OT & $1-t$ & $t$ & ${\bf x}_t - \sigma_t D({\bf x}_t, t; \Theta)$\\
        \bottomrule
    \end{tabular}}
    \caption{Summary of the formula for the estimate of the clean image corresponding to the different formulations. Using the following parameterization, $\forall t, {\bf x}_t = \alpha_t {\bf x}_0 + \sigma_t \pmb{\epsilon}_t$.}
    \label{tab:formulations}
    \vspace{-2em}
\end{SCtable}

\subsection{Generalization for Latent Generative Modeling}

While the bulk of our experiments have been conducted on models trained under DDPM~\citep{ho2020denoising} for noise prediction, we can generalize our method to different frameworks such as diffusion with velocity prediction~\citep{salimans22iclr}  and flow matching~\citep{lipman2023flow}.
To do this, the only requirement is to be able to estimate the original latents from the model predictions.
Under general frameworks such as DDPM  and flows, we can write the forward equation in the form $\forall t, x_t = \alpha_t {\bf x}_0 + \sigma_t \pmb{\epsilon}_t$, where $\alpha_t$ and $\sigma_t$ are increasing (resp. decreasing) functions of $t$.
In \Cref{tab:formulations}, we provide a summary for these different formulations.

\section{Experimental evaluation}

In this section, we first present our experimental setup, and then go on to present our main results, as well as qualitative results and a number of ablation studies.
\vspace{-1em}

\subsection{Experimental setup}

\mypar{Datasets}
We conduct an extensive evaluation on three  datasets of different scales and distributions: 
\imnet~\citep{deng2009imagenet}, 
\cc~\citep{changpinyo2021cc12m},
and \shst: a large internal dataset of 320M stock images.
We note that for both \imnet and \cc, human faces were blurred to avoid training models on identifiable personal data.
For both \cc and \shst, we recaption the images using Florence-2~\citep{xiao24florence2} to obtain captions that more accurately describe the image content.
For each of these datasets, we conduct evaluations at both $256\!\times\!256$ and $512\!\times\!512$ image resolution.

\mypar{Architectures}
All experiments are performed using the Multi-modal DiT architecture from \citet{sd3}.
We downscale the model size to be similar to Pixart-$\alpha$ \citep{chen2023pixartalpha} and DiT-XL/2~\citep{Peebles2022DiT}, which corresponds to $28$ blocks with a hidden size of 1,536, amounting to a total of $796$M parameters.
For \imnet models we condition on class labels, while for the other datasets we condition on text prompts.
For our main results, we perform our experiments using the asymetric autoencoder from~\cite{zhu2023designing}.
For ablation studies, we revert to the lighter autoencoder from SDXL~\citep{podell2024sdxl}.

\mypar{Training and sampling}
Unless specified otherwise, we follow the DDPM-$\epsilon$ training paradigm~\citep{ho2020denoising}, using the DDIM~\citep{song2020denoising} algorithm with $50$ steps for sampling and a classifier-free guidance scale of $\lambda=2.0$~\citep{ho2021classifierfree}.
Following \cite{podell2024sdxl}, we use a quadratic scheduler with $\beta_\text{start}=0.00085$ and $\beta_\text{end}=0.012$.
For the flow experiments, we use the conditional OT probability path~\citep{lipman2023flow} with the mode sampling with heavy tails paradigm from \cite{sd3}.
Under this paradigm, the model is trained for velocity prediction and evaluated using the Euler ODE solver.
For all velocity models, zero terminal SNR schedule is enforced following \cite{lin2024common}.

Similar to~\citet{chen2023pixartalpha}, we pre-train all models at $256$ resolution on the dataset of interest for 600k iterations.
We then enter a second phase of training, in which we optionally apply our perceptual loss, which lasts for  200k  iterations for  $256$ resolution models and for 120k iterations for models at $512$ resolution.

When changing the resolution of the images, the resolution of the latents changes by the same factor, keeping the same noise threshold $\tau_\sigma$ yields inconsistent results across resolutions.
To ensure consistent behavior, we follow \cite{sd3}, and scale the noise threshold similarly to how the noise schedule is scaled in order to keep the same uncertainty per patch.
In practice, this amounts to scaling the threshold by the upscaling factor.
The kernel sizes for the morphological operations in the outlier detection algorithm are also scaled to cover the same proportion of the image.

\mypar{Metrics}
To evaluate our models, we report results  in terms of FID~\citep{heusel17nips} to assess image quality and to what extent the generated images  match the distribution of training images, and CLIPScore~\citep{hessel21emnlp} to assess the alignment between the prompt and the generated image for text-conditioned models.
In addition, we report distributional metrics precision and recall~\citep{precision_recall_distributions} as well as density 
and coverage~\citep{naeem2020reliable} to better understand effects on image quality (precision/density) and diversity (recall/coverage).
We evaluate metrics with respect to \imnet and, for models trained on \cc and \shst, the validation set of \cc. For FID and other distributional metrics, we use the evaluation datasets as the reference datasets and compare an equal number of synthetic samples. For CLIPScore, we use the prompts of the evaluation datasets and the corresponding synthetic samples.
Following previous works \citep{ldm, Peebles2022DiT}, we use a guidance scale of $1.5$ for resolutions of $256$ and $2.0$ for resolutions of $512$, which we also found to be optimal for our baseline models trained without \lpl.

\subsection{Main results}

\begin{figure*}[t]
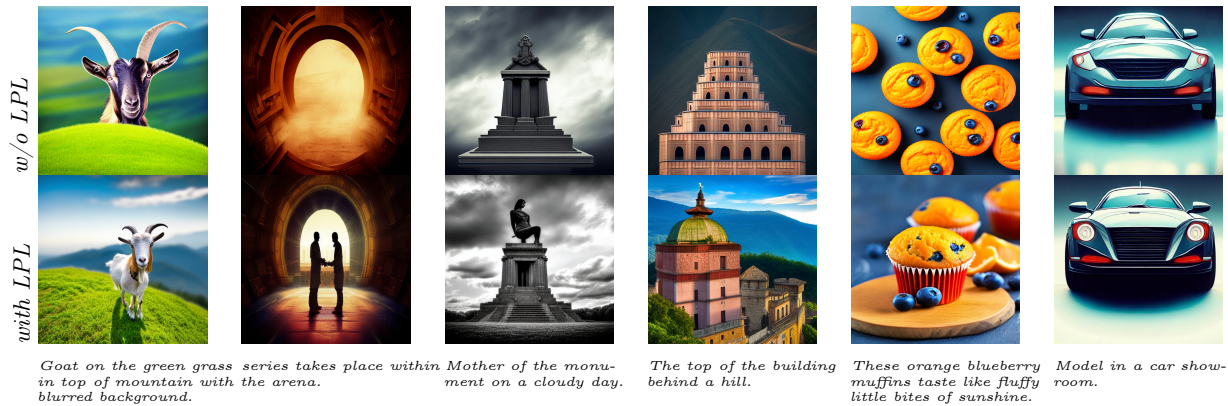

    \def\myim#1#2{\includegraphics[width=22.4mm,height=22.4mm]{figures/sstk_512/v0/#1/#2.png}}
    \tiny
    \centering
    \setlength\tabcolsep{1pt}
    \renewcommand{\arraystretch}{0.2}
    \begin{tabularx}{\linewidth}{cYYYYYY}
     \begin{sideways} \small{\it  w/o \lpl} \end{sideways}    
     & \myim{no_lpl}{000762}
    & \myim{no_lpl}{000223}     
     & \myim{no_lpl}{000880}  
     & \myim{no_lpl}{000995}
     & \myim{no_lpl}{000969}
     & \myim{no_lpl}{000497}
     \\
\begin{sideways} \small{\it  with \lpl} \end{sideways}     
     & \myim{lpl}{000762}
     & \myim{lpl}{000223}     
     & \myim{lpl}{000880}  
     & \myim{lpl}{000995}
     & \myim{lpl}{000969}
     & \myim{lpl}{000497}
     
\\
     & {\it  Goat on the green grass in top of mountain with blurred background.} 
    & {\it  series takes place within the arena.} 
     & {\it  Mother of the monument on a cloudy day.} 
     & {\it  The top of the building behind a hill.} 
     & {\it  These orange blueberry muffins taste like fluffy little bites of sunshine.} 
     & {\it  Model in a car showroom.} 
     \\
    \end{tabularx}
    \caption{
    {\bf Samples from models trained with and without our latent perceptual loss on S320M.} 
    Samples from the model with perceptual loss (bottom row) show more realistic textures and details. 
    }
    \label{fig:lpl_ss}
\end{figure*}

\mypar{\lpl applied across different datasets}
In \Cref{tab:main_results}  we consider the impact of the \lpl on the FID and CLIPScore for models trained on the three datasets and two resolutions. 
We observe that the \lpl loss consistently improves both metrics across all three datasets. 
Most notably, FID is improved by $0.91$ points on \imnet at $512$ resolution and by  $1.52$ points on \cc at 512 resolution.
The CLIPScore is also improved for both resolutions on \cc, by $0.06$ and $0.24$ points respectively.
Similarly, for the \shst dataset, we observe that FID is improved by $0.51$ points while CLIP score improves (marginally) by $0.02$ points.
Samples of models trained with and without \lpl on \cc and \shst are shown in \Cref{fig:lpl_cc12m} and \Cref{fig:lpl_ss}, respectively.

\begin{SCtable}[20]
    \centering
     {\scriptsize
    \begin{tabular}{lrcrcccc}
    \toprule
         & \multicolumn{2}{c}{{\bf Pre-training}} & \multicolumn{3}{c}{{\bf Post-training}} & \multicolumn{2}{c}{{\bf Results}}\\
         \cmidrule(lr){2-3} \cmidrule(lr){4-6} \cmidrule(lr){7-8}
         & {\it Res.} & {\it Iters} & {\it Res.} & {\it Iters} & {\it LPL} & {\it FID $(\downarrow)$} & {\it CLIP $(\uparrow)$}\\
         \midrule
         \multirow{4}{*}{\it \imnet} & \multirow{4}{*}{256} & \multirow{4}{*}{600k} & \multirow{2}{*}{256} & \multirow{2}{*}{200k} & \xmark & $2.98$ & --- \\
         &  & &  & & \cmark & \bf 2.72 & ---\\
         \cmidrule{4-8}
         & & & \multirow{2}{*}{512} & \multirow{2}{*}{120k} & \xmark & $4.88$ & --- \\
         &  & &  & & \cmark & ${\bf 3.79}$ & ---\\
        \midrule
        \multirow{4}{*}{\it \cc} & \multirow{4}{*}{256} & \multirow{4}{*}{600k} & \multirow{2}{*}{256} & \multirow{2}{*}{200k} & \xmark & $7.81$ & $25.06$ \\
         & & & & & \cmark & ${\bf 6.22}$ & ${\bf 25.12}$\\
         \cmidrule{4-8}
         & & & \multirow{2}{*}{512} & \multirow{2}{*}{120k} & \xmark & $8.79$ & $24.88$ \\
         & & & & & \cmark & ${\bf 7.27}$ & ${\bf 25.12}$ \\
         \midrule
         \multirow{2}{*}{\it \shst} & \multirow{2}{*}{256} & \multirow{2}{*}{600k} & \multirow{2}{*}{512} & \multirow{2}{*}{120k} & \xmark & $8.81$ & $24.39$\\
         &  & &  & & \cmark & ${\bf 8.30}$ & ${\bf 24.41}$\\
        \bottomrule
    \end{tabular}}
    \caption{{\bf Impact of our perceptual loss for models trained  on different datasets and resolutions for DDPM-$\epsilon$ model}.
    All models use the same ImageNet-256 pretraining for $600k$ iterations before performing comparing the effect of \lpl during post-training.
    Using \lpl boosts FID and CLIP score for all datasets and  resolutions considered.}
    \label{tab:main_results}
\end{SCtable}

\mypar{Generalization to other frameworks}
We showcase the generality of the \lpl by applying it to different generative models, experimenting with DDPM for both epsilon and velocity prediction, and flow matching with optimal-transport (OT) path similar to \cite{sd3}.
In \Cref{tab:lpl_paradigms} we report experimental results for models trained  on \imnet at 512 resolution.
The DDPM-based models perform very similar (except perhaps for FID, where they differ by 0.16 points), and we we find significant improvements across all metrics other than density when using \lpl.
Density remains similar to the baseline for DDPM models, but improves from 1.14 to 1.29 for the Flow-OT model, where all metrics are improved relative to the DDPM trained ones.
We posit that this is due to the mode sampling scheme in \citep{sd3}, which emphasizes middle timesteps that could better control the trajectories of the flow path towards having more diversity and not improving the quality (precision/density). Hence, applying LPL to Flow-OT solve this by considerably boosting quality.
Notably, considering DDPM baselines, LPL provides a boost as significant as the one provided by using flow matching (scores of DDPM w/ LPL in 2nd and 4th columns are on par or better than Flow-OT w/o LPL in 5th column). Moreover, the provided boost is orthogonal to the training paradigm, leading to overall best results when using \lpl with the flow model.

\begin{SCtable}[50]
    \centering
    {\scriptsize
    \begin{tabular}{lcccccc}
        \toprule
        Paradigm & \multicolumn{2}{c}{DDPM-$\epsilon$} & \multicolumn{2}{c}{DDPM-$v_\Theta$} & \multicolumn{2}{c}{Flow-OT} \\
        \lpl & \xmark & \cmark & \xmark & \cmark & \xmark & \cmark\\
        \midrule
        FID $(\downarrow)$ & $4.88$ & ${\bf 3.79}$ & $4.72$ & ${\bf 3.84}$ & $4.54$ & ${\bf 3.61}$\\
        Coverage $(\uparrow)$ & $0.80$ & ${\bf 0.82}$ & $0.80$ & ${\bf 0.83}$ & $0.82$ & ${\bf 0.85}$\\
        Density $(\uparrow)$ & ${\bf 1.14}$ & $1.13$ & ${\bf 1.15}$ & $1.14$ & $1.14$ & ${\bf 1.29}$\\
        Precision $(\uparrow)$ & $0.74$ & ${\bf 0.77}$ & $0.73$ & ${\bf 0.78}$ & $0.75$ & ${\bf 0.79}$\\
        Recall $(\uparrow)$ & $0.49$ & ${\bf 0.51}$ & $0.49$ & ${\bf 0.50}$ & $0.52$ & ${\bf 0.54}$\\
        \bottomrule
    \end{tabular}}
    \caption{\linespread{1.} {\bf Effect of \lpl on \imnet models at 512 resolution trained with different methods.}
    We observe consistent improvements on all metrics when incorporating the \lpl, except for density metric for which we observe a very slight degradation when using DDPM training.}
    \label{tab:lpl_paradigms}
    \vspace{-2em}
\end{SCtable}

\begin{figure}
     \centering
     \begin{subfigure}[bt]{0.3\textwidth}
         \centering
         \includegraphics[width=.9\textwidth]{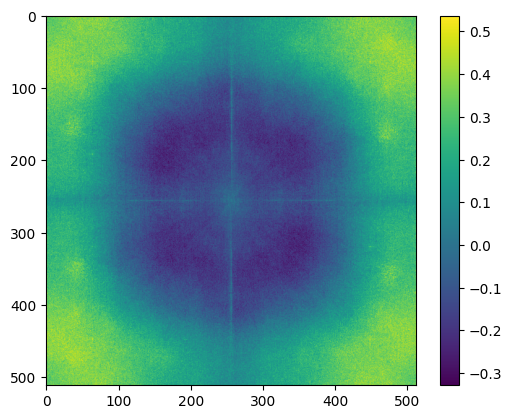}
         \caption{$\mathcal{F}(D_\text{LPL}) - \mathcal{F}(D_\text{base})$}
         \label{fig:lpl_nolpl}
     \end{subfigure}
     \hfill
     \begin{subfigure}[bt]{0.3\textwidth}
         \centering
         \includegraphics[width=.9\textwidth]{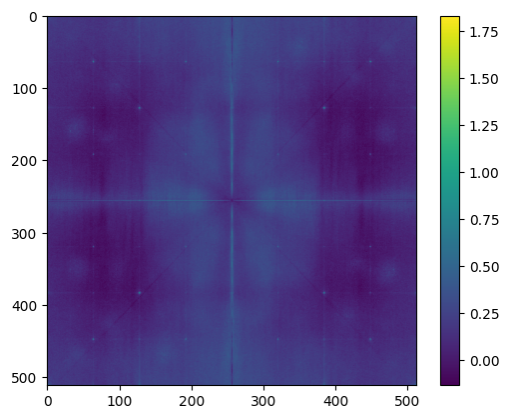}
         \caption{$\mathcal{F}(D_\text{base}) - \mathcal{F}(D_\text{real})$ }
         \label{fig:no_lpl_real}
     \end{subfigure}
     \hfill
     \begin{subfigure}[bt]{0.3\textwidth}
         \centering
         \includegraphics[width=.9\textwidth]{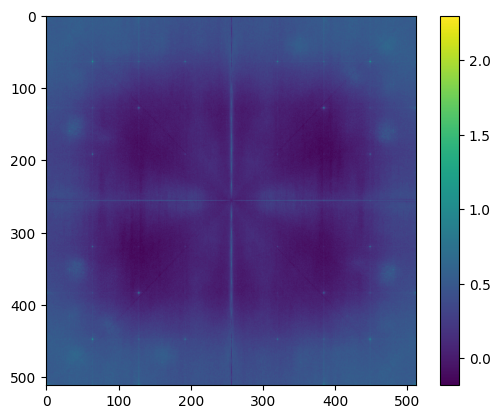}
         \caption{$\mathcal{F}(D_\text{LPL}) - \mathcal{F}(D_\text{real})$}
         \label{fig:lpl_real}
     \end{subfigure}
        \caption{{\bf Power spectrum of real and generated images.} Difference in (log) power spectrum  between image generated with and without LPL. Using LPL strenghtens frequencies at the extremes (very low and very high). 
        }
        \label{fig:power_spectrums}
        \vspace{-2em}
\end{figure}

\mypar{Frequency analysis}
While the metrics above provide useful information on model performance, they do not specifically provide insights in terms of frequencies at which using LPL is more effective at modeling data than the baseline. 
To provide insight on the effect of our perceptual loss \wrt the frequency content of the generated images, we compare the power spectrum profile of images generated with a model trained with and without LPL on \cc at 512 resolution as well as a set of real images from the validation set.

\begin{wrapfigure}{r}{.4\linewidth}
    \vspace{-1em}
    \centering
    \includegraphics[width=.9\linewidth]{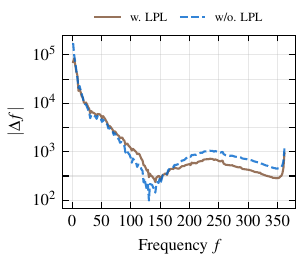}
    \caption{{\bf Frequency comparison.} We compare the power spectrum of the images obtained with or without LPL with real reference images from the validation set of CC12M.
    }
    \label{fig:freq_comp}
\end{wrapfigure}
In \Cref{fig:power_spectrums}, we plot the difference between log-power spectra between the three image sets.
The left-most panel clearly shows  the presence of more high frequency signal in the generated images when using LPL to train the model, confirming what has been observed in the qualitative examples of \Cref{fig:lpl_cc12m} and \Cref{fig:lpl_ss}.
Moreover, the very lowest frequencies are also strengthened in the samples of the model with \lpl. 
We posit that using the LPL makes it easier to match very low frequencies as they tend be encoded separately in certain channels of the decoder.
In \Cref{fig:freq_comp}, we report the error when comparing the power-spectrum of synthetic images and real images, averaged across the validation set of \cc.
For this, we compute the average of the power spectrum across a set of 10k synthetic images from each model and the reference images for the validation set of \cc.
Our experiments indicate that the model trained with \lpl is consistently more accurate in modeling high frequencies ($f>150$), at the expense of a somewhat larger error at middle frequencies ($75<f<150$).

\subsection{Ablation results}

\begin{figure}[t]
\vspace{-1em}
    \centering
    \begin{minipage}[t]{0.3\textwidth}
        \centering
        \includegraphics[width=\linewidth]{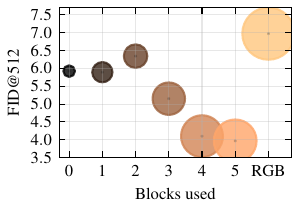}
        \caption{\linespread{1.} \small {\bf Exploration of \lpl depth.} Influence of decoder blocks used in \lpl on FID, 0 corresponds to not using \lpl. 
        Disk radius shows GPU memory usage:  w/o LPL=64.9 GB, \lpl-5 blocks=83.4 GB. 
        }
      \label{fig:lpl_layers}
    \end{minipage}
    \hspace{0.01\textwidth}
    \begin{minipage}[t]{0.3\textwidth}
        \centering
        \includegraphics[width=\linewidth]{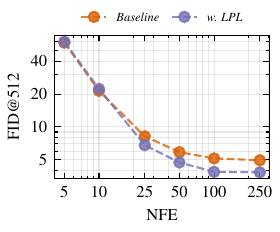}
        \caption{\small {\bf Impact of \lpl for different number of sampling steps.} With higher numbers of sampling steps, the difference between the baseline and the model trained with \lpl increases.
        }
        \label{fig:lpl_convergence}
    \end{minipage}
        \hspace{0.01\textwidth}
        \begin{minipage}[t]{.3\textwidth}
        \centering
        \includegraphics[width=\linewidth]{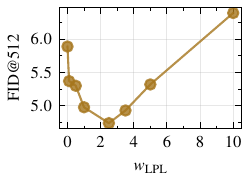}
         \caption{\small {\bf Influence of the \lpl loss weight on model performance.} The curve shows a sharp decrease in FID before going back up for larger weights.
         }
         \label{fig:lpl_w}
    \end{minipage}%
    \vspace{-1em}
\end{figure}

\mypar{\lpl depth}
Using  decoder layers  to compute our perceptual loss comes with increased computational and memory costs.
We therefore  study the effect of computing our perceptual loss using only a subset of the decoder layers, as well as a baseline using the RGB pixel output of the decoder.
We progressively add more decoder features, so the model with five blocks  contains features from the first up to the fifth block. 
From the results in \Cref{fig:lpl_layers}, we find that earlier blocks do not significantly improve FID - and can even negatively impact performance. 
Deeper layers, on the other hand, significantly improve the performance. 
The most significant gains are obtained when incorporating the third and fourth decoder layers which both improve the FID by more than one point \wrt the model incorporating one block less.
Finally, the last decoder layer improves the FID only marginally and could be omitted to reduce resource consumption. 
We perform an additional ablation where the loss operates directly on the RGB image space (without using internal decoder features), which results in degraded performance compared to the baseline while inducing a considerable memory overhead.

\mypar{Feature normalization}
Before computing our perceptual loss, we normalize the decoder features. 
We compared normalizing the features of the original latent and the predicted one separately, or normalizing both using the statistics from the predicted latent. Our experiment is conducted on \imnet at 512 resolution. 
While the model trained with separately normalized latents results in a slight boost of FID ($4.79$ \vs  $4.88$ for the baseline w/o \lpl), the model trained with shared normalization statistics leads to a much more significant improvement and obtains an FID of $3.79$.

\mypar{SNR threshold value}
We conduct an experiment on the influence of the SNR threshold which determines at which time steps our perceptual loss is used for training. 
Lower threshold values correspond to using \lpl for fewer iterations that are closest to the noise-free targets.
We report results across several metrics in \Cref{fig:sigma_ths} and illustrated with qualitative examples in the supplementary material in \Cref{fig:noise_threshold}.
We find improved performance over the baseline without \lpl  for all metrics and that the best  values for each metric are obtained for a threshold between three and six, except for the recall which is very stable (and better than the baseline) for all threshold values under 20.

\begin{figure}[t]
    \centering
        \vspace{-2mm}
    \includegraphics[width=\linewidth]{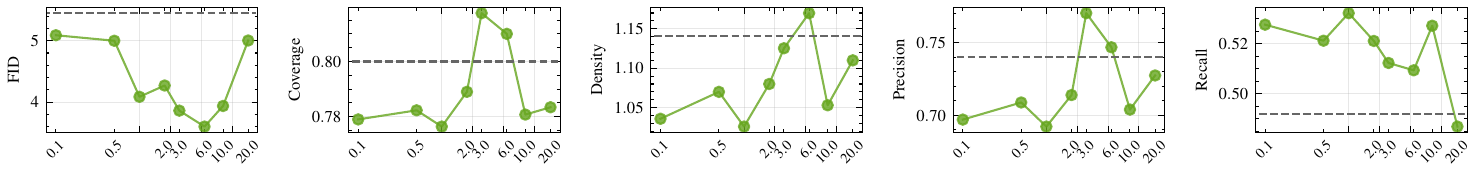}
    \caption{{\bf Ablation study on the impact of the noise  threshold $\tau_\sigma$.} 
    We report FID, coverage, density, precision and recall.
    The dashed line corresponds to the baseline without LPL, 
    note the logarithmic scaling of the noise threshold on the horizontal axis.
    }
    \label{fig:sigma_ths}
    \vspace{-2mm}
\end{figure}

\mypar{Reweighting strategy}
We compare the performance when using uniform or depth-specific weights to combine the contributions from different decoder layers in the \lpl. 
We find that using depth-specific weights results in significant improvements in terms of image quality \wrt using uniform weights. While the depth-specific weights achieve an FID of $3.79$, the FID obtained using uniform weights is $4.38$.
Hence, while both strategies improve image quality over the baseline (which achieves an FID of $4.88$), reweighting the layer contributions to be approximately similar further boosts performance and improves FID by $0.59$ points.

\mypar{\lpl and convergence}
As the \lpl loss adds a non-negligible memory overhead, by having to backpropagate through the latent decoder, it is interesting to explore  at which  point in training it should be introduced.
We train models on \imnet at 512 resolution with different durations of the post-training stage.
We use an initial post-training phase --- of zero, 50k, or 400k iterations --- in which \lpl is not used, followed by another 120k iterations in which we either apply \lpl or not.

The results in  \Cref{tab:lpl_convergence} indicate that in each case \lpl improves all metrics and that the improvements are larger  when the model has been trained longer and is closer to convergence (except for the coverage metric where we see the largest improvement when post-training for only 120k iterations).
This suggests that better models (ones trained for longer) benefit more from our perceptual loss.

\begin{SCtable}[50]
    \centering
    {\scriptsize
    \begin{tabular}{lccccc}
        \toprule
        Initial post-train iters & 0 & 50k  & 400k\\
        \midrule
        $\Delta$ FID $(\downarrow)$& $-0.58$ & $-0.78$  & $-0.97$\\
        $\Delta$ coverage $(\uparrow)$ & $+4.29$ & $+3.51$ & $+3.99$ \\
        $\Delta$ density $(\uparrow)$ & $+0.14$ & $+0.12$ & $+0.21$\\
        $\Delta$ precision $(\uparrow)$ & $+4.01$ & $+4.55$ & $+5.89$\\
        $\Delta$ recall $(\uparrow)$ & $+1.99$ & $+2.32$ & $+4.22$\\
        \bottomrule
    \end{tabular}}
    \caption{
    {\bf Effect of our perceptual loss on models pre-trained without \lpl for a set number of iterations.} 
    In each column, we report the difference in metrics  after post-training for 120k iterations with or without \lpl. 
    All metrics improve when adding \lpl in the post-training phase.
    }
    \label{tab:lpl_convergence}
    \vspace{-1.5em}

\end{SCtable}

\mypar{Influence on sampling efficiency}
We conduct an experiment to assess the influence of the perceptual loss on the sampling efficiency.
To this end, we sample the ImageNet@512 model with different numbers of function evaluations (NFE) then check the trends for the baseline and the model trained with our method.
For this experiment, we use DDIM algorithm.
Results are reported on \Cref{fig:lpl_convergence}, where we find that for very low numbers of function evaluations, both models perform similarly. 
The improvement gains from the \lpl loss start becoming considerable after $25$ NFEs, where we observe a steady increase in performance gains with respect to the number of function evaluations up to $100$, afterwards both models stabilize at a point where the model trained using \lpl achieves an improvement of approximately $1.1$ points over the baseline.

\begin{wrapfigure}{r}{0.4\textwidth}
    \vspace{-1em}
    \centering
    \includegraphics[width=.9\linewidth]{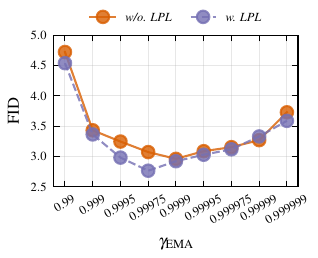}
    \caption{{\bf Impact of EMA decay rate.} 
    Training with \lpl is more stable, and allows for a smaller decay parameter.
    }
    \label{fig:ema_influence}
\end{wrapfigure}
\mypar{Impact on EMA}
Since the \lpl has the effect of increasing the accuracy of the estimated latent during every timestep, it reduces fluctuations between successive iterations of the model during training.
Consequently, when training with \lpl the EMA momentum can be reduced to obtain  optimal performance.
In \Cref{fig:ema_influence} we report the results of  a grid search over the momentum parameter $\gamma_\textrm{EMA}$.
We find that the model trained with \lpl achieves better results when using a slightly lower momentum than the baseline.
From the graph, it's clear that better FID is obtained closer to $\gamma_\text{EMA} = 0.99975$ for the \lpl model, which corresponds to a half-life of approximately $2750$ iterations, while the non-\lpl model achieves its optimal score at $\gamma_\text{EMA} = 0.9999$ corresponding to a half life of $6930$ iterations, more than twice as much as the \lpl model, thereby validating our hypothesis.

\mypar{Relative weight}
We conduct a grid search over different values for the weight of the \lpl loss $w_\text{LPL}$. We report FID after training for 120k iterations at $512$ resolution, all models are initialized from the same $256$ pretrained checkpoint.
Our results are reported in \Cref{fig:lpl_w}.
Introducing \lpl sharply decreases FID for lower weights before going back up at higher weights.
We find the model to achieve the best FID for $w_{\lpl} \approx 3.0$ which roughly corresponds to a fifth of the relative contribution to the total loss.

\section{Conclusion}
In this work, we identified a disconnect between the decoder and the training of latent diffusion models, 
where the diffusion model loss does not receive any feedback from the decoder resulting in perceptually non-optimal generations that oftentimes lack high frequency details. 
To alleviate this disconnect we introduced a latent perceptual loss (LPL) that provides perceptual feedback from the autoencoder's decoder when training the generative model. Our quantitative results showed that the LPL is generalizable and improves performance for models trained on a variety of datasets, image resolutions, as well as generative model formulations. 
We observe that our loss leads to improvements from 6\% up to 20\% in terms of FID. 
Our qualitative analysis show that the introduction of LPL  leads to models that produce images with better structural consistency and sharper details compared to the baseline training. 
Given its generality, we hope that our work will play an important role in improving the quality of future latent generative models.

\newpage
\bibliography{iclr2025_conference}
\bibliographystyle{iclr2025_conference}

\appendix
\newpage

\section{Appendix}

\subsection{Relevance of the LPL loss} \label{app:lpl_rel}

\mypar{LPL as an image-space projection penalty}
In the following, we interpret LPL in the simple setting where the decoder is approximated by a linear mapping $F_\beta^d ({\bf z}) = \hat{ \bf x} =  {\bf A} {\bf z}$.

Under the DDPM paradigm, for latent diffusion, a neural network is trained to model the reverse process $q({\bf z}_{t-1}|{\bf z}_t)$.
Under this setting, training is conducted by optimizing the KL divergence between the true reverse process and the predictor that is modeled using a neural network:
\begin{equation}
    \mathcal{L}_{t-1} = \mathbb{E}_q \left[ D_\text{KL} \left( q({\bf z}_{t-1}|{\bf z}_t, {\bf z}_0) \parallel p_\Theta({\bf z}_{t-1} | {\bf z}_t) \right) \right].
\end{equation}

Taking into account the global objective which is image generation, we argue that the accuracy of the sampling steps should be measured in image space and not in latent space, this means {\it putting more emphasis on obtaining $\ell_2$-optimal reconstructions in image space rather than in latent space}.
Such a constraint can be imposed in the form of a penalty term that is added to the training objective:
\begin{equation} \label{eq:penalty}
    \mathcal{L}_{t-1}^\text{pen} = \mathbb{E}_q \left[ D_\text{KL} \left( q({{ \bf x}}_{t-1}|{{ \bf x}}_t, {{ \bf x}}_0) \parallel p_\Theta({\hat{ \bf x}}_{t-1} | {\hat{ \bf x}}_t) \right) \right].
\end{equation}

Following DDPM~\cite{ho2020denoising} notation, the ground truth and predicted forward process posterior distributions are given by:
\begin{align}
    q({{ \bf x}}_{t-1}|{{ \bf x}}_t, {{ \bf x}}_0) & = \mathcal{N} \left( {\bf x}_{t-1}; \tilde{\pmb \mu}_t \left( {\bf x}_t, {\bf x}_0 \right)
, \tilde{\beta}_t {\bf I} \right)\\
    p_\Theta({\hat{ \bf x}}_{t-1} | {\hat{ \bf x}}_t) & = \mathcal{N} \left( \hat{\bf x}_{t-1}; \pmb{\mu}_\Theta (x_t, t), \sigma^2 {\bf I} \right).
\end{align}
In this formula, we can introduce the linear mapping to observe the latent variables instead.
\begin{align}
    q({{ \bf x}}_{t-1}|{{ \bf x}}_t, {{ \bf x}}_0) & = \mathcal{N} \left( {\bf x}_{t-1}; {\bf A} \tilde{\pmb \mu}_t ({\bf z_t}, {\bf z}_0), \tilde{\beta} {\bf A} {\bf A}^{\top} \right)\\
    p_\Theta({\hat{ \bf x}}_{t-1} | {\hat{ \bf x}}_t) & = \mathcal{N} \left(\hat{{\bf x}}_{t-1}; {\bf A} {\pmb \mu}_\Theta({\bf z}_t, t), \sigma^2 {\bf A} {\bf A}^{\top}  \right).
\end{align}

Subsequently, \cref{eq:penalty} can be developed using the closed form of KL divergence between two Gaussian distributions:

\begin{align}
    \mathcal{L}_{t-1}^\text{pen} & = \frac{1}{2} \mathbb{E}_q \left[ \log \frac{\lvert \sigma^2 \rvert}{\lvert \tilde{\beta}_t \rvert} + ({\bf A}( \tilde{\pmb \mu}_t - \pmb{\mu}_\Theta))^\top \sigma^{-2} \left( {\bf A} {\bf A}^\top \right)^{-1} ({\bf A} ( \tilde{\pmb \mu}_t - \pmb{\mu}_\Theta)) + \left( \frac{\sigma^2}{\tilde{\beta}_t} -1 \right)\text{Tr} \left\{ {\bf I} \right\}  \right]\\
    \mathcal{L}_{t-1}^\text{pen} & = \frac{1}{2\sigma^2} \mathbb{E}_q \left[ \big( {\bf A}( \tilde{\pmb \mu}_t - {\pmb \mu}_\Theta) \big)^\top \left( {\bf A} {\bf A}^\top \right)^{-1} \big( {\bf A} ( \tilde{\pmb \mu}_t - {\pmb\mu}_\Theta) \big) \right] + \text{C}.
\end{align}

An upper bound for this term can be obtained by taking the largest eigenvalue of the pseudo-inverse $\left( {\bf A} {\bf A}^\top \right)^{-1}$.
\begin{equation}
    \mathcal{L}_{t-1}^\text{pen} \leq \frac{1}{2\sigma^2} \lambda_{\max} \left( \left( {\bf A} {\bf A}^\top \right)^{-1} \right)  \mathbb{E}_q \left[ \lVert {\bf A} ( \tilde{\pmb \mu}_t - {\pmb\mu}_\Theta) \rVert_2^2 \right] + \text{C}.
\end{equation}

From this equation, the image-level penalty can be interpreted as equivalent to optimizing the reconstruction between the real image and the decoded latent prediction $\lVert {\bf A} \hat{\bf z}_0 ({\bf z}_t, t; \Theta) - {\bf x}_0 \rVert_2^2$.

In the more general case, where the decoder is not a linear mapping, we can use a first order Taylor expansion to obtain a linear approximation, assuming that ${\bf z}_0$ and $\hat{{\bf z}}_0$ are close enough, which is reasonable for earlier timesteps.
\begin{equation}
    \hat{\bf z}_0 ({\bf z}_t, t; \Theta) = D({\bf z}_t, t; \Theta) = {\bf z}_0 + \sigma_{{\bf z}_t} {\bf n},
\end{equation}
where $\sigma_{{\bf z}_t} {\bf n}$ is small compared to ${\bf z}_0$.
\begin{equation}
    \hat{\bf x}_0 ({\bf z}_t, t; \Theta) = F^d_\beta(\hat{\bf z}_0) = {\bf x}_0 + \sigma_{{\bf z}_t} \nabla_{{\bf z}_0} F^d_\beta ({\bf z})^\top {\bf n}.
\end{equation}
Which brings us back to the linear case that was tackled above.
Consequently, for earlier timesteps, \lpl loss can be interpreted as an image space penalty that pushes the forward process posteriors to be more accurate in images space rather than in latent space.

\subsection{Latent Structure}
\begin{figure}[ht]
    \centering
    \includegraphics[width=\linewidth]{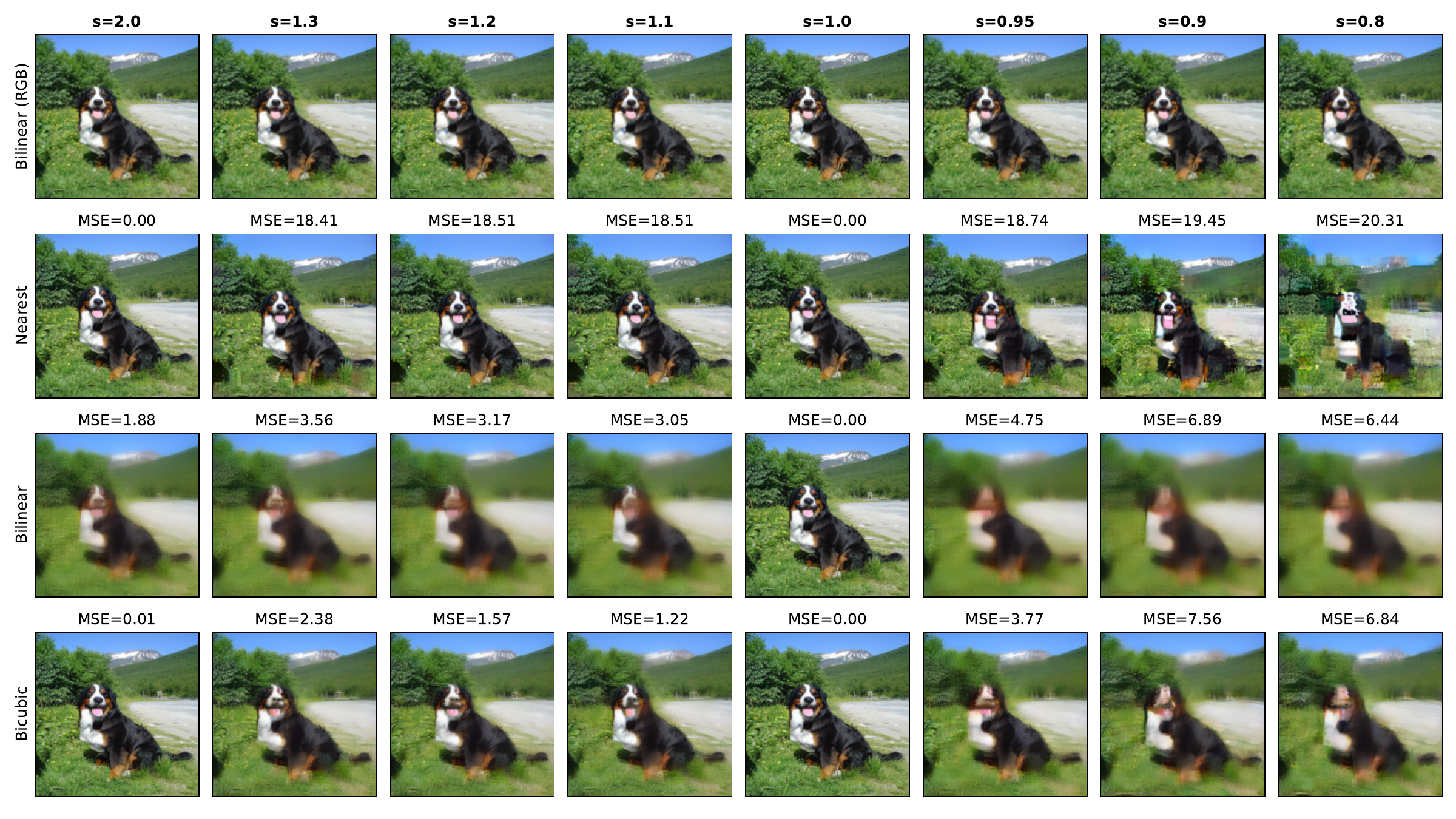}
    \caption{{\bf Influence of interpolation artefacts on latent reconstruction.} We downscale the image by a factor of $1/s$ before upscaling back to recover the original resolution. 
    {\it From top to bottom:} bilinear interpolation in pixel space, nearest in latent space, bilinear in latent space and bicubic interpolation in latent space.
    }
    \label{fig:interp_pb}
    \vspace{-1em}
\end{figure}

Because of the underlying structure of the latent space, certain errors can have much more detrimental effects to the quality of the decoded image than others.
We illustrate this in \Cref{fig:interp_pb} by comparing the generated image after interpolating the encoded latents to different resolutions then back to its original resolution before decoding them.
While these different transformations yield similar errors in terms of MSE, especially in RGB space, the interpolation algorithm becomes crucial when working in the latent space.

An illustration of this effect is presented in \Cref{fig:interp_pb} where we degrade the quality of the latents by performing an interpolation operation to downsize the latents (when $s<1$) followed by the reverse operation to recover latents at the original size, such a transformation can be seen as a form of lossy compression where different interpolation methods induce different biases in the information lost.

By examining the reconstructions from the latents, we cannot conclude that there is a direct relationship between the MSE with respect to the original latent and the decoded image quality.
While nearest interpolation results in the highest MSE, the reconstructed images are more perceptually similar to the target than the ones obtained with bilinear interpolation.
Similarly, while the bicubic interpolation with $s=1.3$ achieves an MSE  of  $2.38$, it still results in better reconstruction than the bilinear interpolation where $s=2.0$ which achieves a lower MSE error of $1.88$.

From this analysis, we see that certain kinds of errors can have more or less detrimental effects on the image generation, which go beyond simple MSE in the latent space.

\begin{figure}
    \centering
    \includegraphics[width=\linewidth]{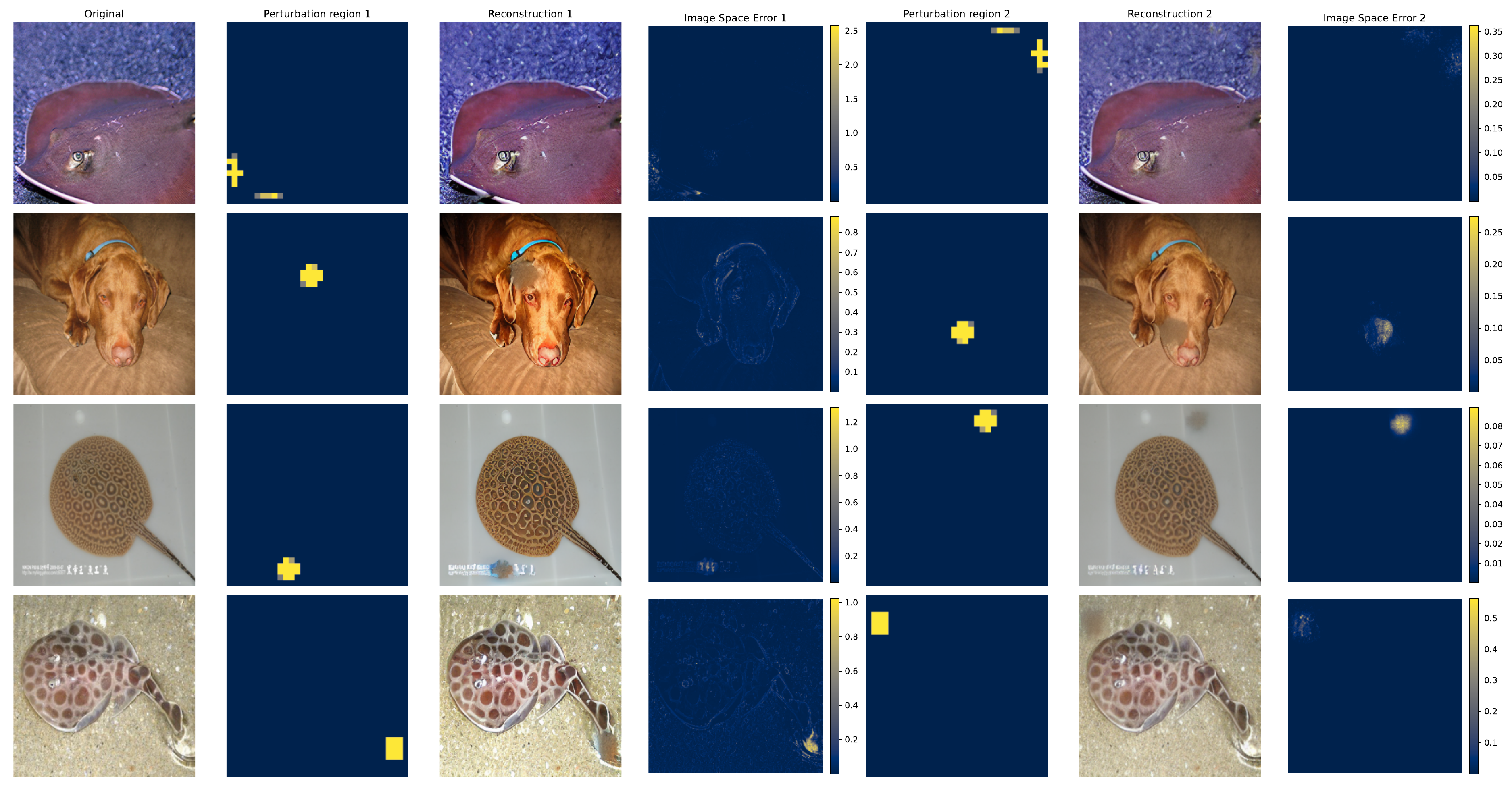}
    \caption{{\bf Illustration of the irregularity of the SD AE.} Certain regions of the image induce a global error in the image and higher error norm, while the same perturbation in other locations in the image results in much lower and localized error.}
    \label{fig:vae_irregularity}
\end{figure}

Another experiment illustrating the irregularity of the autoencoder's latent space is given in \Cref{fig:vae_irregularity}.
In this experiment, we select certain pixel in latent space to which we add a slight random noise (that is half the variance of the latents), we afterwards reconstruct the image from the perturbed latent.
What we can see is that depending on the perturbed region, the error in image space can differ significantly.
Most notably, we uncover many cases where masking a region in latent space can degrade the reconstruction quality over the whole image.
This also showcases that certain high-level information can be contained in certain spatial regions of the latents while others play a relatively less important role.

\subsection{Outlier Detection}

At deeper layers of the autoencoder, some layers have aretfacts where small patches in the feature maps have a norm orders of magnitude higher than the rest of the feature map.
These aretefacts have been detected consistently when testing the different opensource autoencoders available online, which include the ones used in our experiments\footnote{ 
\url{https://huggingface.co/stabilityai/sdxl-vae}, and
\url{https://huggingface.co/cross-attention/asymmetric-autoencoder-kl-x-1-5}}, 
as well as others.\footnote{
\url{https://huggingface.co/CompVis/stable-diffusion-v1-4}, and 
\url{https://huggingface.co/cross-attention/asymmetric-autoencoder-kl-x-2}.
}

\begin{figure}[t]
    \centering
    \includegraphics[width=.9\linewidth]{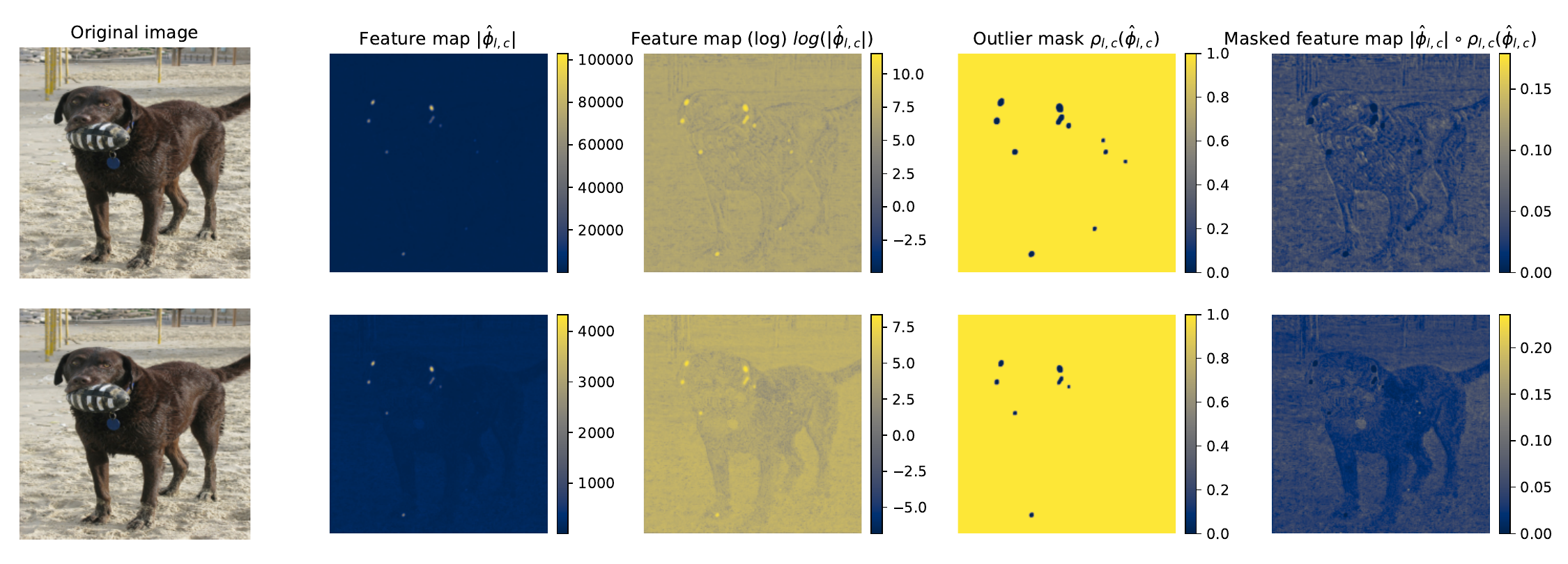}
    \caption{{\bf Example of feature maps from the autoencoder's decoder.} The presence of outliers makes the underlying feature representation difficult to exploit.
    $l$ refers to the block index, while $c$ is the channel index within the block.
    Top row: $l=4,c=2$, bottom row: $l=4,c=8$. 
    }
    \label{fig:outlier}
\end{figure}

\mypar{Outlier detection}
When inspecting the decoder features  we find artefacts at decoder's deeper layers. Particularly, in some cases a small number of decoder activations have very high  absolute values, see \Cref{fig:outlier}.
This is  undesirable, as such outliers can dominate the perceptual loss, reducing its effectiveness. 
To prevent this, we use a simple outlier detection algorithm to mask them when computing the perceptual loss.
See the supplementary material for details. 

To ensure easy adaptability to different models, we propose a simple detection algorithm for these patches and mask them when computing the loss and normalizing the feature maps.
Our algorithm is based on simple heuristics and is not meant to provide a state-of-the-art solution for outlier detection.
Rather, it is proposed as a temporary patch for the observed issues, while the long-term solution would be to train better autoencoders that do not suffer from these outliers.

\mypar{Detection algorithm}
We empirically observe that the activations for every feature map  approximately follow a normal distribution, while the outliers can be identified as a small subset of out-of-distribution points.
To identify them, we threshold the points with the corresponding percentile at $\delta_o$ and $1-\delta_o$ percentiles.
Since computing the quantiles can be computationally expensive during training, we do it using nearest interpolation, which amounts to finding the $k$-th largest value in every feature map where $k = \delta_o \times H_f \times W_f$ (or $k = (1 - \delta_o) \times H_f \times W_f$ for the maximal values).
To remove small false positives that persist in the outlier mask, we apply a morphological opening, which can be seen as an erosion followed by a dilation of the feature map.
Pseudo-code for the outlier detection algorithm is provided in Alg.\ \ref{lst:outlier_det}.

\begin{figure}[t]
\renewcommand{\lstlistingname}{Algorithm}
\lstset{
  language=Python,
  basicstyle=\small\fontfamily{zi4}\selectfont,
  keywordstyle=\color{blue},
  commentstyle=\color{green!50!black},
  stringstyle=\color{red!50!black},
  numbers=none,
  frame=single,
  rulecolor=\color{gray!20},
  backgroundcolor=\color{gray!10},
  captionpos=b,
}
    
\begin{lstlisting}[caption={{\bf Outlier detection algorithm.} The algorithm works by setting a threshold according to the upper $0.02$ quantile of the activations in the feature map. Because the outliers are orders of magnitude away from the rest, we shift the threshold by an offset $m$ that guarantees that only the outliers are thresholded while no activations are masked when no outliers are present. Subsequently, we smooth out the predicted mask using a dilation operation that eliminates small noise in the mask.}, label={lst:outlier_det}]
def remove_outliers(features, down_f=1, opening=5,
                        closing=3, m=100, quant=0.02):
    opening = int(ceil(opening/down_f))
    closing = int(ceil(closing/down_f))
    
    if opening == 2:
        opening = 3
    if closing ==2:
        closing = 1
    
    # replace quantile with kth (nearest interpolation).
    feat_flat = features.flatten(-2, -1)
    
    k1 = int(feat_flat.shape[-1]*quant)
    k2 = int(feat_flat.shape[-1]*(1-quant))
    
    q1 = feat_flat.kthvalue(k1, dim=-1).values[..., None, None]
    q2 = feat_flat.kthvalue(k2, dim=-1).values[..., None, None]

    # Mask obtained by thresholding at the upper quantiles.
    m = 2*feat_flat.std(-1)[..., None, None].detach()
    mask = (q1-m < features)*(features < q2+m)
    
    # dilate the mask.
    mask=MaxPool2d(
            kernel_size=closing, 
            stride=1,
            padding=(closing-1)//2
        )(mask.float()) # closing
        
    mask=(-MaxPool2d(
            kernel_size=opening,
            stride=1,
            padding=(opening-1)//2
        )(-mask)).bool() # opening
        
    features = features * mask
    return mask, features
\end{lstlisting}
\end{figure}

\subsection{Perceptual Losses in Diffusion}
In this section, we provide a comparison to recent related works that incorporate some form of perceptual objective into diffusion model training. 

\mypar{E-LatentLPIPS}
\cite{kang2024diffusion2gan}, train a classifier in the autoencoder latent space to define an LPIPS metric. 
While their models are only tested for distilling diffusion models into GANs, we experiment with using them as training losses and compare their performance with \lpl.
To this end, we only include the $\ell_2$ objective and do not apply any augmentations to the inputs for the loss.

\mypar{Self-Perceptual}
\cite{lin2024diffusionmodelperceptualloss}, develop a Self-Perceptual loss where the intermediate features of the denoiser network are used as basis for the perceptual loss that is used during training.
this loss is not directly applicable in our use-case as it was developed on a UNet model, which shows a significantly different structure from the state-of-the-art DiT networks.
Most notably, the UNet encoder progressively downsamples the feature maps before upsampling them again to their original size, corresponding blocks of the same resolution are linked with skip connections.
Hence, the deepest encoder block results in a semantic map description that can be obtained relatively efficiently.
On the other hand, for DiT-type models, the successive blocks all share the same resolution, making it unclear as to which block is optimal for this use-case.
We note, however, that \citet{lin2024diffusionmodelperceptualloss} report best performance when using the standard $\ell_2$ loss  with classifier-free guidance, \ie better than when using their Self-Perceptual loss. 
Therefore its practical usefulness remains unclear.
To evaluate this method with state-of-the-art DiT-type architectures, we conduct experiments where the loss is computed at different depths of the network and compare the results with our LPL.

\mypar{Comparison}
To ensure a fair comparison, we set the weight of the loss such as the variance of the perceptual loss is $0.1$ of the variance of the $\ell_2$ loss when $\ell_2$ loss is also part of the training objective.
Results are summarized in \Cref{tab:perceptual_losses}.

In terms of {\it training throughput} (using the asymmetric auteoncoder at $512$ resolution), we observe a decrease of approximately $42\%$ when training with \lpl \vs when not training with it, this is significant but is still manageable considering it only applies at later training stages.
In comparison, Self-Perceptual with the middle layer applied to MMDiT reduces throughput by $31\%$, such effect should get larger with larger models, making the use of this method prohibitive for very large transformer architecture, especially if used in conjunction with EMA, in which case three different copies of the model should be kept in memory.
E-LatentLPIPS on the other hand is more efficient ($20\%$ throughput reductions) as it uses lightweight models that operate at low resolutions.\\
In terms of {\it performance}, none of the models that we compare with achieve better FID scores than the baseline.
For Self-Perceptual, we observe the highest degradation when using the mid-layer while shallower layers perform the best, although the impact on training performance is negative when compared to the $\ell_2$ baseline.
For E-LatentLPIPS, we also observe a degradation of FID by almost $3$ points when comparing with the baseline.

\begin{table}
    \centering
    {\scriptsize
    \begin{tabular}{lccc}
        \toprule
        Loss & FID $(\downarrow)$ & throughput (img/s) $(\uparrow)$ & \% Mem/GPU $(\downarrow)$\\
        \midrule
        $\ell_2$ & $4.88$ & $16.6$ & $42.5$\\
        \midrule
        E-LatentLPIPS & $7.71$ & $13.3$ & $67.5$\\
        E-LatentLPIPS$^\dag$ & $7.19$ & $15.8$ & $43.5$\\
        \midrule
        Self-Perceptual$_{8}$ & $5.82$ & $13.3$ & $70.6$\\
        Self-Perceptual$_{16}$ & $13.40$ & $11.4$ & $75.1$\\
        Self-Perceptual$_{28}$ & $10.43$ & $9.2$ & $82.4$\\
        \midrule
        LPL & $ 3.79$ & $9.6$ & $74.3$\\
        \bottomrule
    \end{tabular}}
    \caption{
    {\bf Comparison with different perceptual losses in latent space.} For Self-Perceptual, we note the depth at which the loss is computed as a subscript. $\dag$ corresponds to a perceptual loss that is only applied for later timesteps, similar to \lpl. We report FID training throughput and memory usage per GPU (in \%) for the same batch size.
    Throughput and memory measurements are taken during the post-training stage, \ie with perceptual losses applied in all iterations.
    }
    \label{tab:perceptual_losses}
\end{table}

\subsection{Indirect effects of \lpl}
One possible explanation for the improved performance is that it comes from a certain timestep-specific reweighting which puts more emphasis on later timesteps, as samples in the \lpl timestep range  tend to be present both in both the vanilla $\ell_2$ loss and our \lpl.
To verify this claim, we compare the performance with a model trained with a timestep specific reweighting that equalizes the contributions of the timesteps across the batch, the re-weighting is applied as follows:
\begin{equation} \label{eq:rew}
    w(\sigma_t) = 1 + w_\text{LPL} \cdot \frac{\sigma^2_{\mathcal{L}_\text{LPL}}}{\sigma^2_{\ell_2}} \cdot \delta_{\sigma_t \leq \tau_\sigma}(\sigma_t),
\end{equation}
where $\sigma^2_{\mathcal{L}_\text{LPL}}$ (resp. $\sigma^2_{\ell_2}$) is the mean variance of the LPL loss (resp. the vanilla $\ell_2$ objective).
Such a weighting effectively amplifies later timesteps to have the same contribution when using LPL or not using it.
Additionally, we compare the standard epsilon weights with state-of-the-art reweighting strategies such as min-SNR~\citep{Hang_2023_ICCV}, results are reported in \cref{tab:time_reweighting}.
\begin{SCtable}[50]
    \centering
    {\scriptsize
    \begin{tabular}{lc}
        \toprule
        Re-weighting & FID $(\downarrow)$\\
        \midrule
        Baseline & $4.88$\\
        Baseline$^\dag$ & $5.43$\\
        min-SNR & $5.04$\\
        LPL & $3.79$\\
        \bottomrule
    \end{tabular}
    }
    \caption{{\bf Influence of timestep reweighting strategy.} Compared to different timestep reweighting strategies, LPL finetuning achieves significant improvements while timestep reweightings result in degraded performance compared to the baseline epsilon weighting. $\dag$ corresponds to the reweighting from \cref{eq:rew}.}
    \label{tab:time_reweighting}
\end{SCtable}

As seen in \Cref{tab:time_reweighting}, introducing timestep reweighting results in degraded performance compared to the baseline while \lpl significantly improves FID.
Hence the improvements cannot be attributed to time-specific reweighting.

\subsection{Memory overhead}
For a more fair and accurate comparisons with the baseline, we conduct additional experiments where we equalize not the number of iterations but the maximum achievable batch size and the training time, in order to obtain a more accurate sense of the applicability of our loss in practice.\\

\begin{SCtable}
    \centering
    {\scriptsize
    \begin{tabular}{lcccc}
        \toprule
        Method & Batch Size/GPU & (T)FLOP/it & Training iterations (k) & FID $(\downarrow)$\\
        \midrule
        Baseline & $25$ & $108.347$ & $122.5$ & $4.62$\\
        with LPL & $16$ & $106.581$ & $82.5$ & ${\bf 3.84}$\\
        \bottomrule
    \end{tabular}}
    \caption{{\bf Comparisons for the same time budget.} All models are trained with maxed-out batch size for $48$ hours using two A100 nodes per experiment.}
    \label{tab:time_comparison}
\end{SCtable}

\mypar{Memory maximization}
\Cref{tab:time_comparison} compares models trained on the same resources for an equal time duration.
This experiment is conducted on ImageNet@512, both models are trained on 2 A100 nodes for a duration of $48$ hours.
For the baseline with a higher batch size, the learning rate is scaled linearly in order to account for this discrepancy.
Under this setting, the baseline trains with a batch size/GPU of $25$ for $122.5k$ iterations, achieving an FID of $4.62$ while the model with \lpl is trains for $82.5k$ iterations with a batch size/GPU of $16$, achieving an FID of $3.84$.
When comparing FLOPs, we found both runs to have similar FLOPs per iteration.
Hence, while the FID gap between the two models is reduced from $1.09$ to $0.78$, it remains significant.

\begin{figure}
    \centering
    \begin{minipage}[t]{0.45\textwidth}
        \centering
        \includegraphics[width=\linewidth]{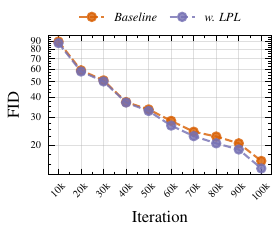}
        \caption{\linespread{1.} \small {\bf \lpl from scratch.} Comparison on the effect of \lpl when applied from scratch.
        }
      \label{fig:compare_scratch}
    \end{minipage}
    \hspace{0.01\textwidth}
    \begin{minipage}[t]{.45\textwidth}
        \centering
        \includegraphics[width=\linewidth]{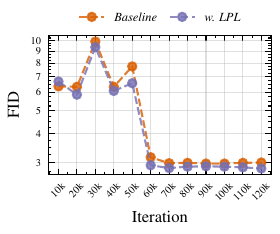}
         \caption{\small {\bf \lpl finetuning.} Comparison on the effect of \lpl when applied as finetuning.
         }
         \label{fig:compare_finetune}
    \end{minipage}
        \vspace{-2mm}
\end{figure}

\begin{figure}
    \centering
    \begin{minipage}[t]{0.31\textwidth}
        \centering
        \includegraphics[width=\linewidth]{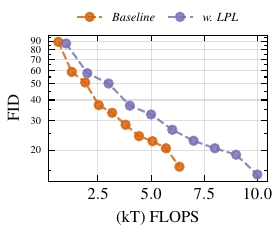}
        \caption{\linespread{1.} \small {\bf \lpl from scratch.} Comparison on the effect of \lpl when applied from scratch.
        }
      \label{fig:compare_scratch}
    \end{minipage}
    \hspace{0.01\textwidth}
    \begin{minipage}[t]{.31\textwidth}
        \centering
        \includegraphics[width=\linewidth]{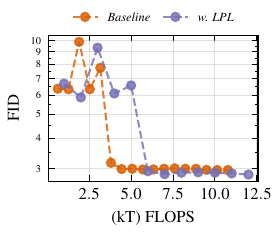}
         \caption{\small {\bf \lpl finetuning.} Comparison on the effect of \lpl when applied as finetuning.
         }
         \label{fig:compare_finetune}
    \end{minipage}%
    \hspace{0.01\textwidth}
    \begin{minipage}[t]{.31\textwidth}
        \centering
        \includegraphics[width=\linewidth]{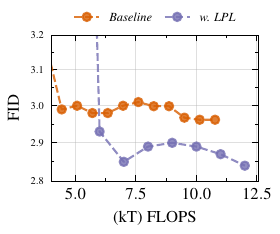}
         \caption{\small {\bf \lpl finetuning.} Crop around EMA region.
         }
         \label{fig:compare_finetune}
    \end{minipage}%
        \vspace{-2mm}
\end{figure}

\mypar{Minimum iterations}
To quantify the needed number of training iterations, we perform finetuning experiments and track the FID every 10k iterations, allowing us to track how long the \lpl should be applied to see benefits. 
These experiments are conducted on \imnet at 256 resolution.

In \Cref{fig:compare_scratch}, we experiment with the effect of training from scratch when using \lpl.
In the beginning of training, both models have similar performance, however as training continues and the performance gain due to LPL increases. 

In \Cref{fig:compare_finetune}, we experiment with finetuning a pre-trained model and track FID every $10k$ iterations. 
The big FID improvement at $50k$ iterations is due to the activation of EMA.
We observe that as of $20k$ iterations, the model with LPL consistently obtains better performance.

\mypar{FLOPS Analysis}
In \Cref{tab:flop_comp}, we conduct a detailed analysis of the FLOP count for the duration of training under different settings, reporting the total FLOP count for different training settings.
We observe that adding \lpl to the training losses increase the FLOPs/iteration by a faactor of approximately $1.5$.
However, when accounting for the number of training iterations, we found the total FLOP increase to be around $10-15\%$.

\begin{table}
    \centering
     {\scriptsize
    \begin{tabular}{lrccrcccccc}
    \toprule
         & \multicolumn{3}{c}{{\bf Pre-training}} & \multicolumn{4}{c}{{\bf Post-training}} & \multicolumn{2}{c}{{\bf Total}}\\
         \cmidrule(lr){2-4} \cmidrule(lr){5-8} \cmidrule(lr){9-10}
         & {\it Res.} & {\it Iters} & {\it (T)FLOPs/it} & {\it Res.} & {\it Iters} & {\it LPL} & {\it (T)FLOP/it} & {\it (T)FLOPS} & FLOPs Increase\\
         \midrule
         \multirow{4}{*}{\it \imnet} & \multirow{4}{*}{256} & \multirow{4}{*}{600k} & \multirow{2}{*}{$63.389$} & \multirow{2}{*}{256} & \multirow{2}{*}{200k} & \xmark & $63.342$ & $50,711.20$k & \multirow{2}{*}{$14.43\%$}\\
         & & & & & & \cmark & $99.987$ & $58,030.80$k & \\
         \cmidrule{4-10}
         & & & \multirow{2}{*}{$63.389$} & \multirow{2}{*}{512} & \multirow{2}{*}{120k} & \xmark & $69.342$ & $46,354.44$k & \multirow{2}{*}{$9.64 \%$}\\
         &  & &  & & & \cmark & $106.581$ & $50,823.12$k & \\
        \midrule
        \multirow{4}{*}{\it \cc/\shst} & \multirow{4}{*}{256} & \multirow{4}{*}{600k} & \multirow{2}{*}{$63.391$} & \multirow{2}{*}{256} & \multirow{2}{*}{200k} & \xmark & $63.391$ & $50,712.80$k & \multirow{2}{*}{$14.46\%$}\\
         & & & & & & \cmark & $100.064$ & $58,047.40$k & \\
         \cmidrule{4-10}
         & & & \multirow{2}{*}{$63.391$} & \multirow{2}{*}{512} & \multirow{2}{*}{120k} & \xmark & $69.342$ & $46,355.64$k & \multirow{2}{*}{$9.64\%$}\\
         & & & & & & \cmark & $106.583$ & $50,824.56$k & \\
        \bottomrule
    \end{tabular}}
    \caption{{\bf FLOP analysis}. We compare the total FLOP increase throughout training under the different settings in our main results table.
    }
    \label{tab:flop_comp}
\end{table}

\subsection{Comparison of baseline to the state of the art}
Our baseline setting uses the DDPM paradigm  with the state-of-the-art  MMDiT architecture  from~\cite{sd3}.
For the Flow-OT model, we follow the same setting as SD3.
To verify that this baseline is competitive with state-of-the-art models, we provide quantitative comparison of our baseline with other open-source models trained on ImageNet at 512 resolution.
Accordingly, we follow the same procedure for the models and report FID for the training set of ImageNet using $250$ sampling steps for each sample.
Our flow model is sampled using an ODE sampler based on the RK4 solver from \cite{torchdiffeq}.
The results in \Cref{tab:sota_baselines} show that 
our flow baseline with RK4 solver achieves state-of-the-art results compared to the results reported in the literature.

\begin{SCtable}
    \centering
    {\scriptsize
    \begin{tabular}{lc}
        \toprule
        Model & FID $(\downarrow)$\\
        \midrule
        UNet~\cite{ldm} & 4.81\\
        DiT-XL/2~\cite{Peebles2022DiT} & 3.04\\
        mDT-v2~\cite{gao2023masked} & 3.75\\
        SiT-XL/2 (Flow-SDE)~\cite{ma2024sitexploringflowdiffusionbased} & 2.62 \\
        \midrule
        mmDiT-XL/2 (DDPM - DDIM) & 3.02\\
        mmDiT-XL/2 (Flow-OT - ODE) & \bf 2.49\\
        \bottomrule
    \end{tabular}}
    \caption{{\bf Comparison with other baselines.} We compare our baseline training with other models in the literature, our baseline training results in state-of-the art performance. FID scores for each model are taken from their respective papers.}
    \label{tab:sota_baselines}
\end{SCtable}

\begin{SCfigure}[0.5][h!]
    \centering
    \includegraphics[width=.45\linewidth]{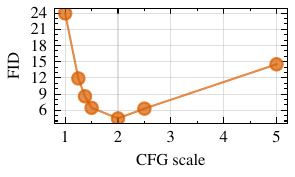}
    \caption{{\bf Influence of guidance scale on FID.} For ImageNet@512, the baseline modela achieves the best FID score with an guidance scale of $w_\text{CFG} = 2.0$.}
    \label{fig:cfg_scale}
\end{SCfigure}
\mypar{Guidance scale}
In the \Cref{fig:cfg_scale}, we report the FID of the baseline model trained on ImageNet@512 for different guidance scale values. We find the best FID to be achieved for $w_\text{CFG} = 2.0$, which is the value chosen in our experiments.

\subsection{Additional qualitative results}

\begin{figure*}
    \def\myim#1{\includegraphics[width=37mm,height=37mm]{figures/threshold_example/threshold_ex#1.png}}
    \centering
    \setlength\tabcolsep{0.5pt}
    \renewcommand{\arraystretch}{0.2}
    \begin{tabular}{ccccc}  
        \begin{sideways}$\tau_\sigma = 0.8$ \end{sideways} & \myim{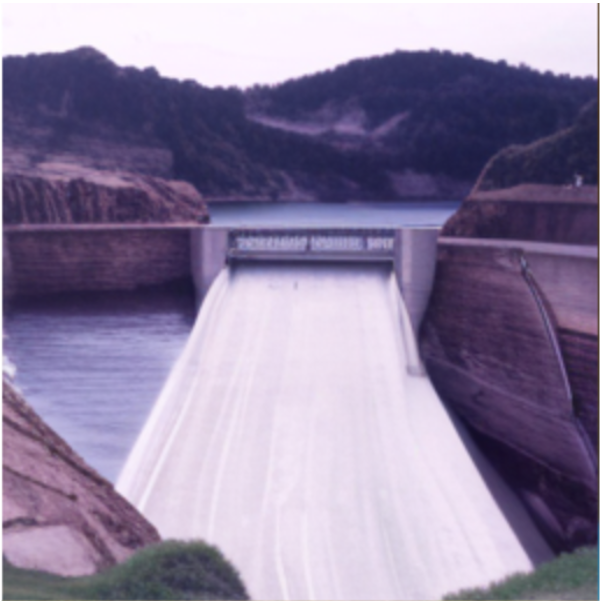} & \myim{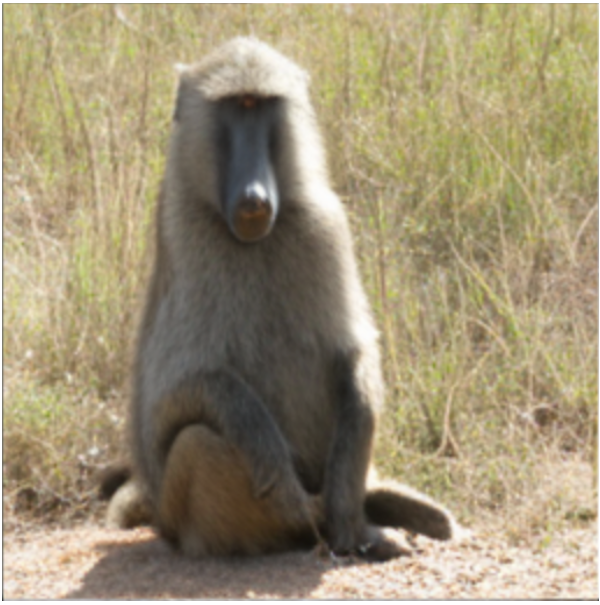} & \myim{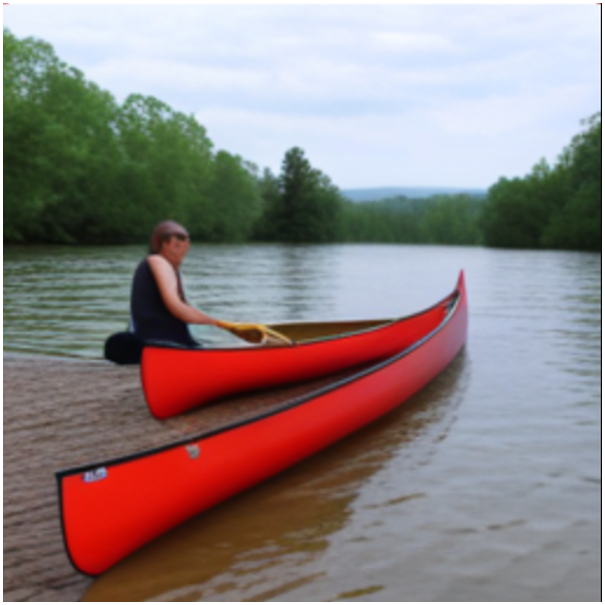} & \myim{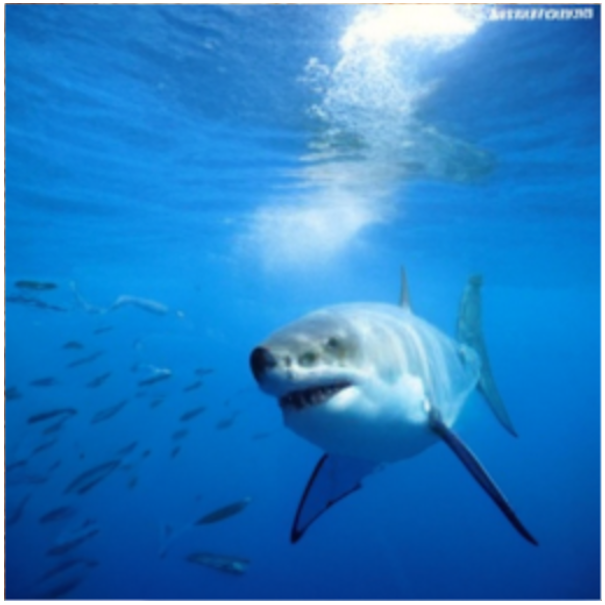}\\
        \begin{sideways}$\tau_\sigma = 2.0$ \end{sideways} & \myim{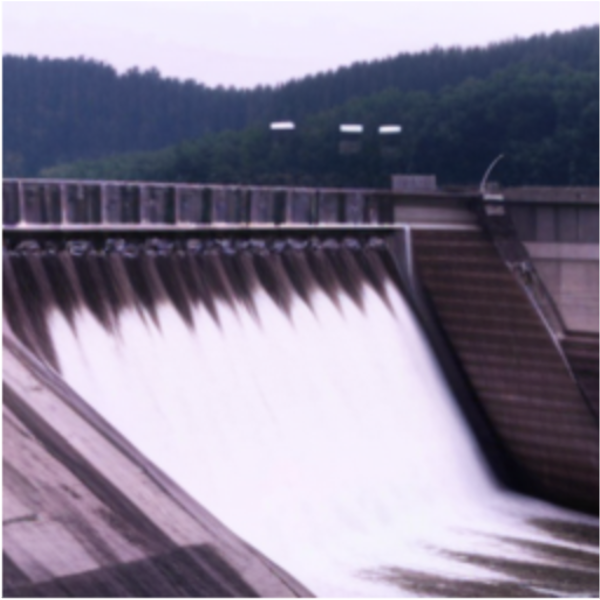} & \myim{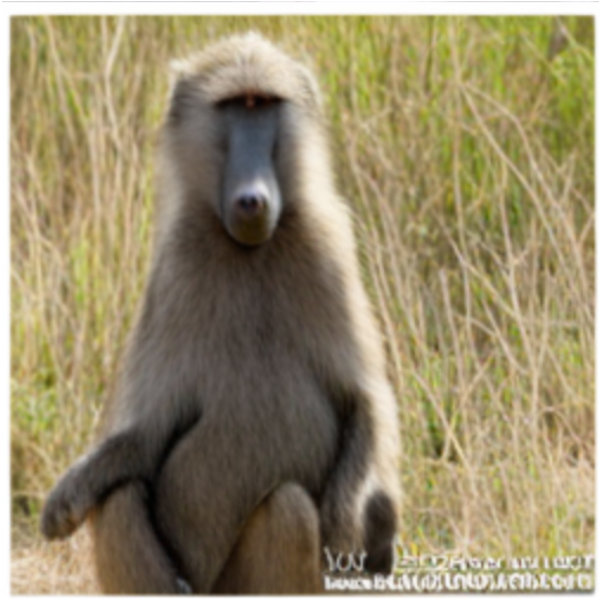} & \myim{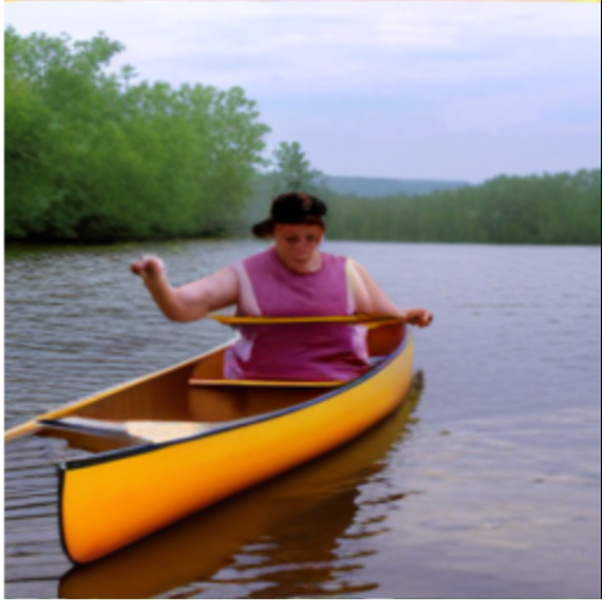} & \myim{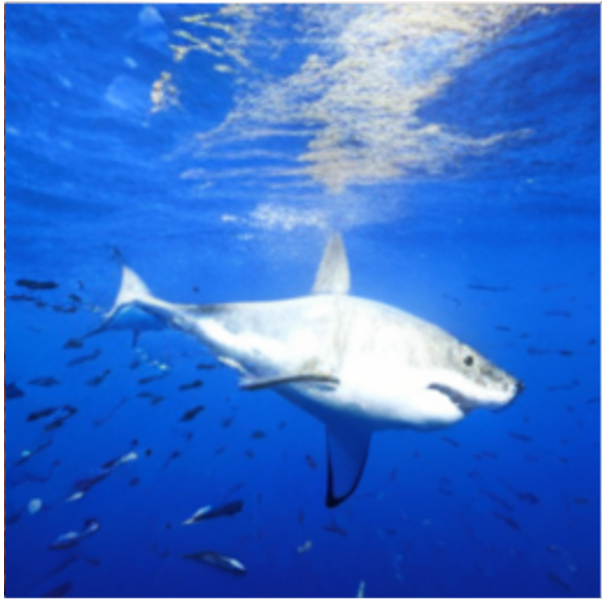}\\
        \begin{sideways}$\tau_\sigma = 3.0$ \end{sideways} & \myim{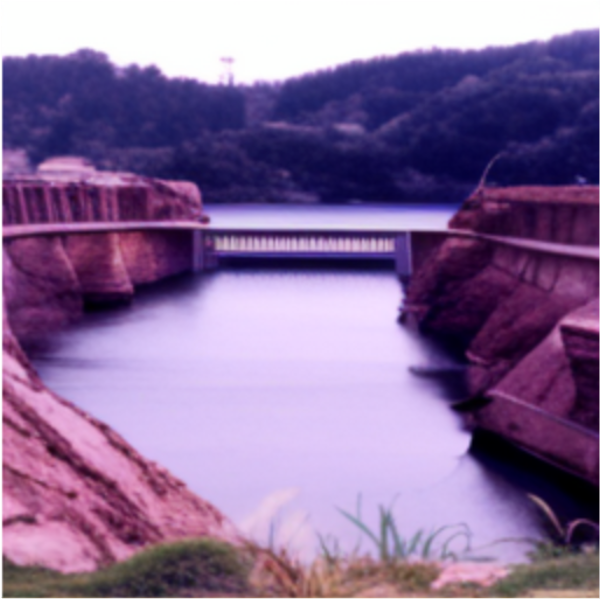} & \myim{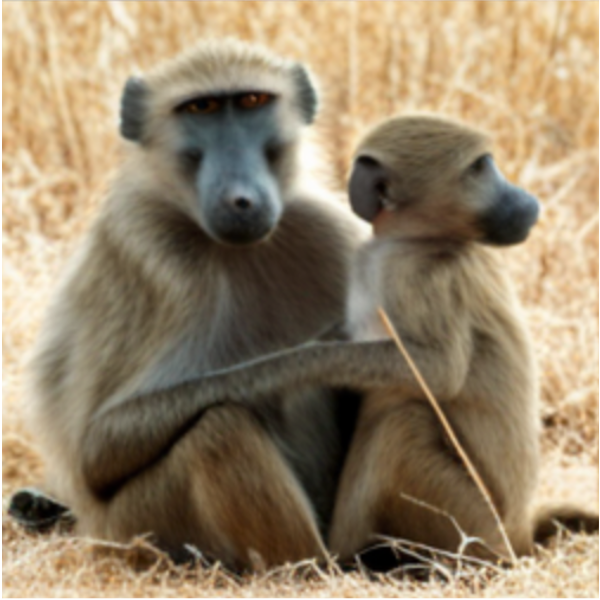} & \myim{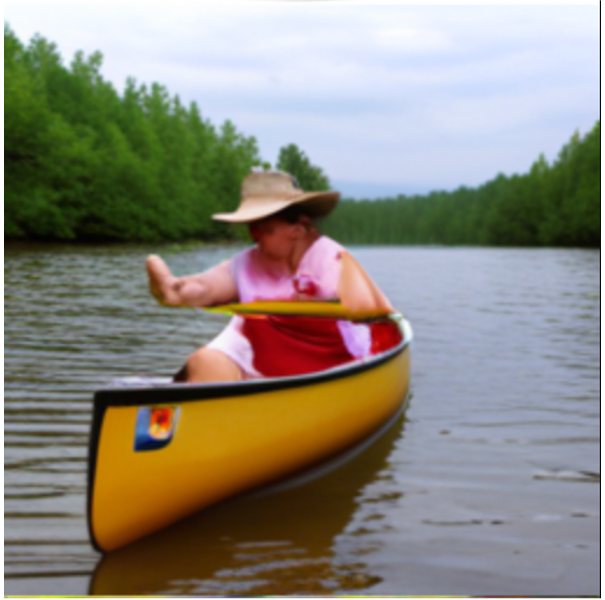} & \myim{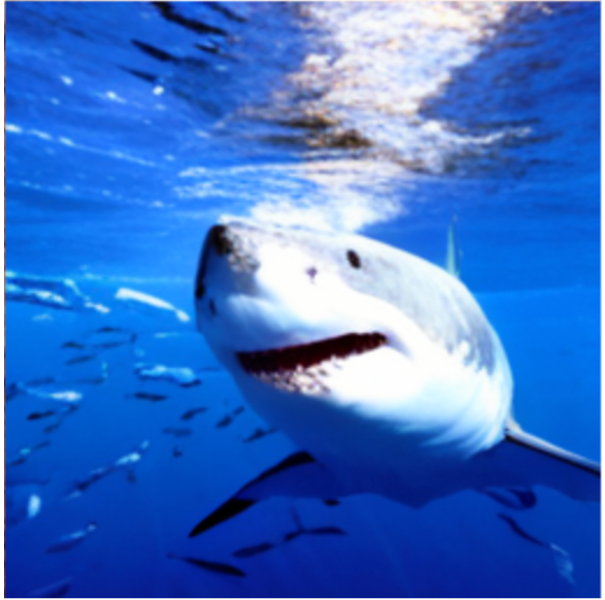}\\
        \begin{sideways}$\tau_\sigma = 0.8$ \end{sideways} & \myim{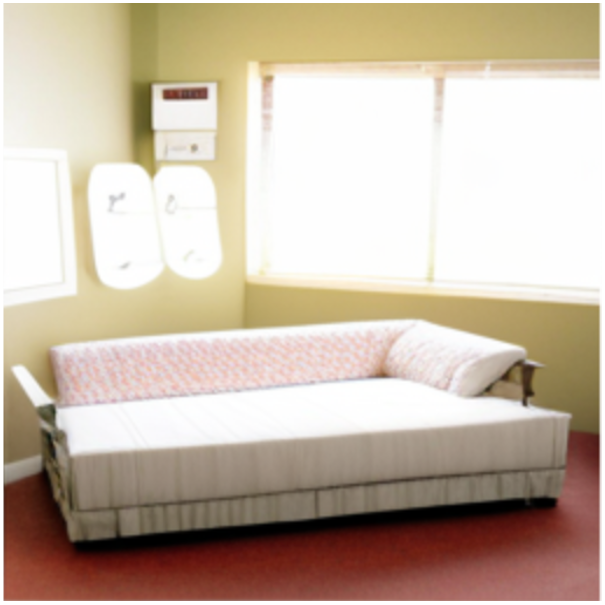} & \myim{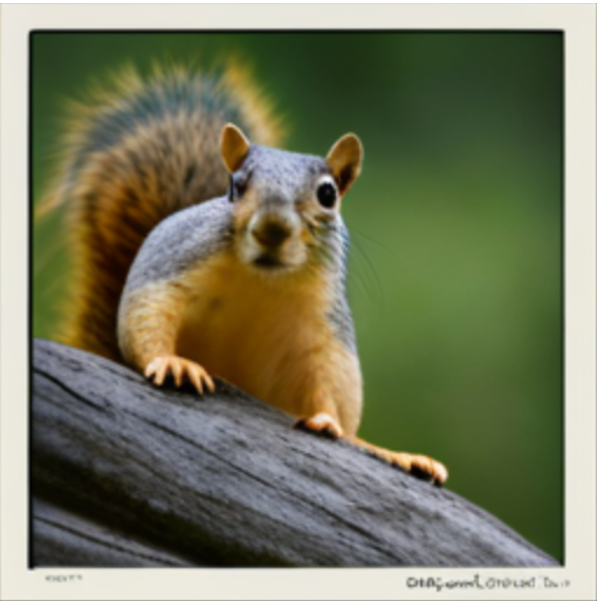} & \myim{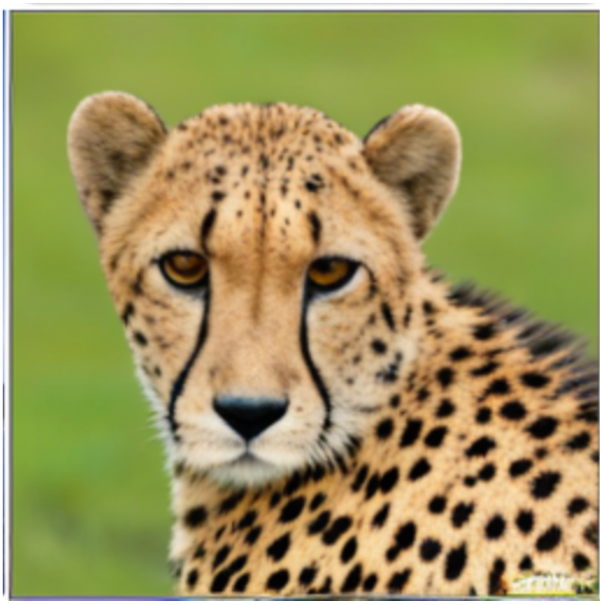} & \myim{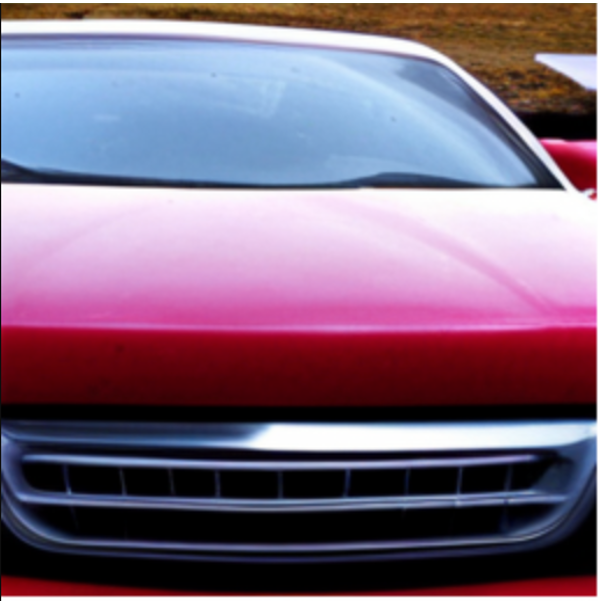}\\
        \begin{sideways}$\tau_\sigma = 2.0$ \end{sideways} & \myim{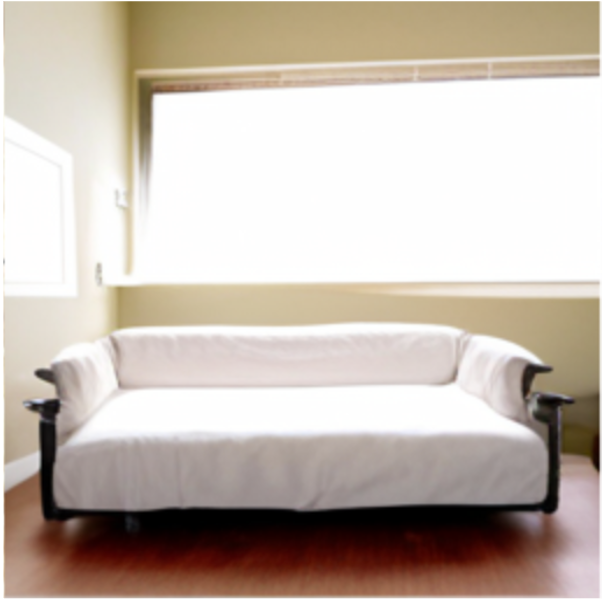} & \myim{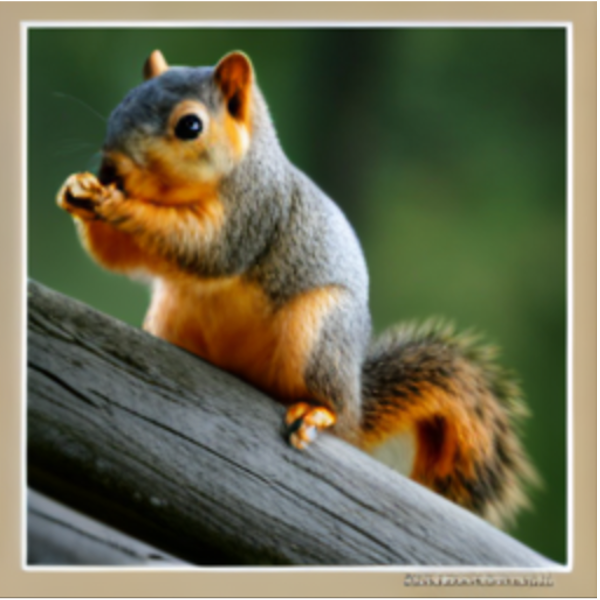} & \myim{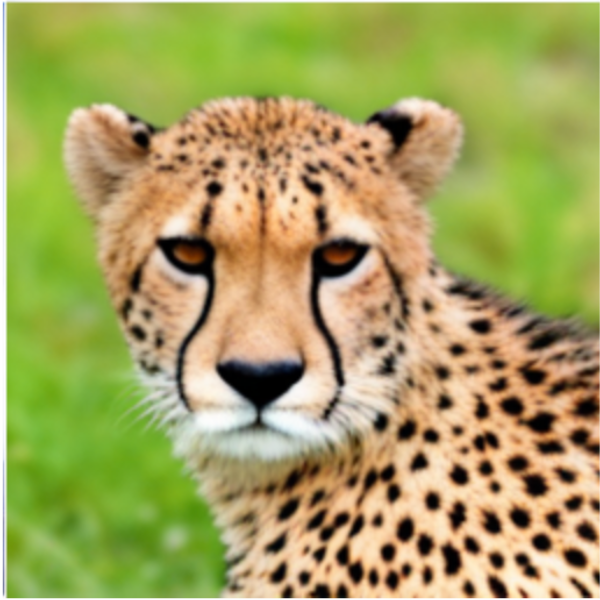} & \myim{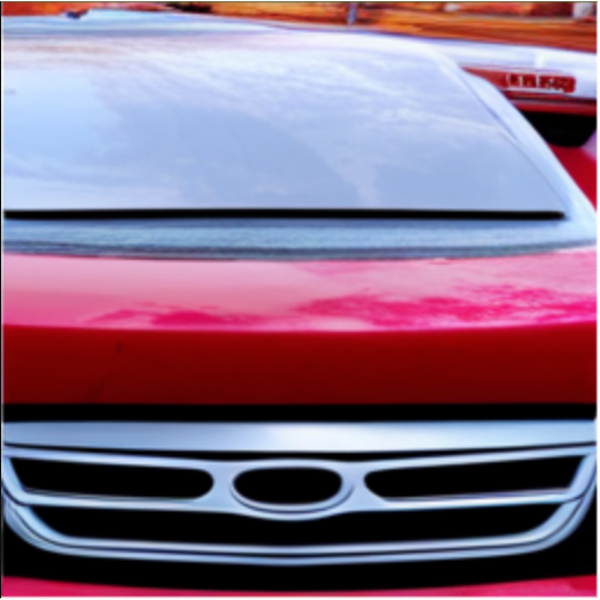}\\
        \begin{sideways}$\tau_\sigma = 3.0$ \end{sideways} & \myim{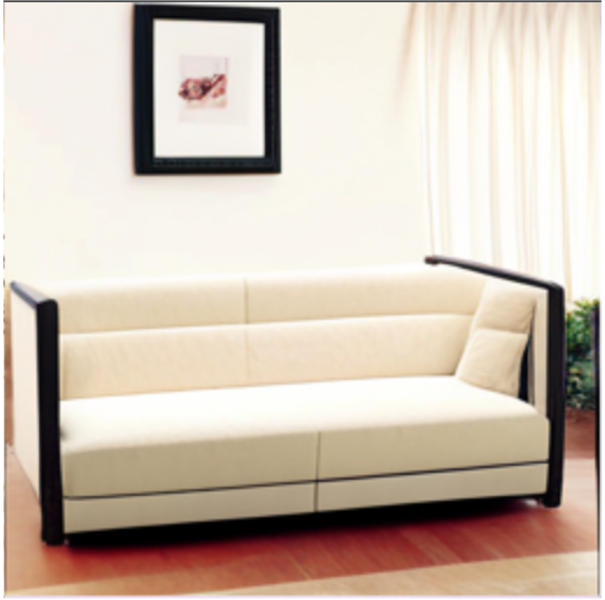} & \myim{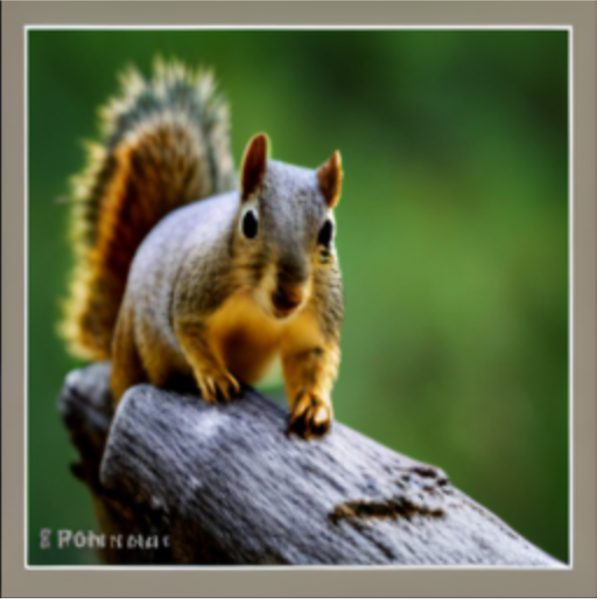} & \myim{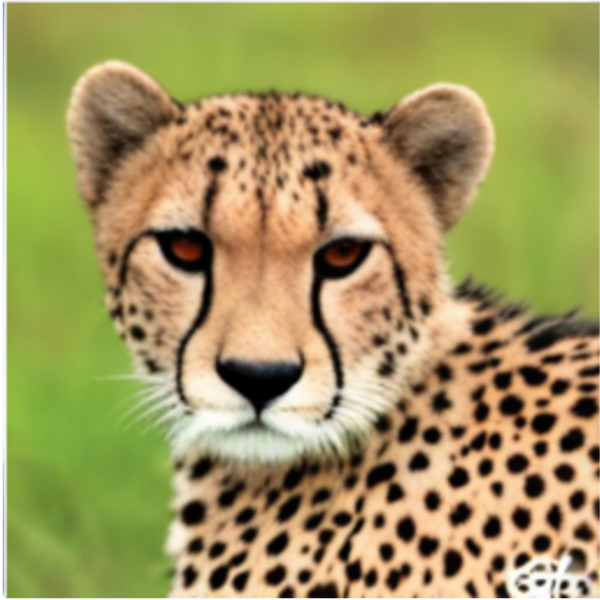} & \myim{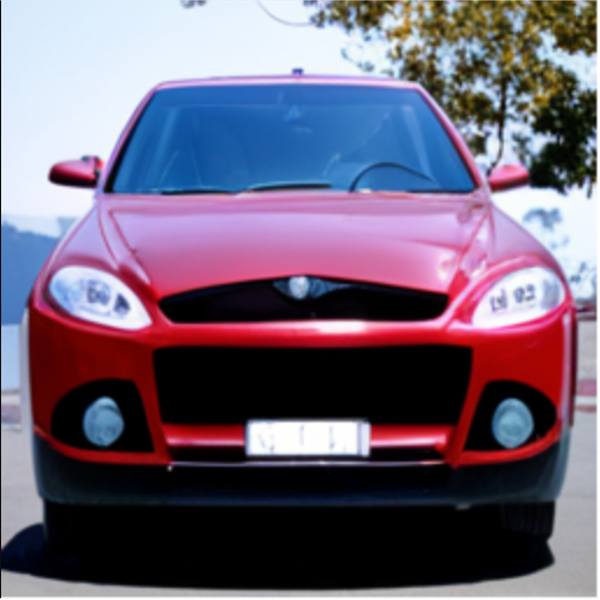}\\
    \end{tabular}
    \caption{{\bf Influence of noise threshold.} Higher thresholds allow for more detailed and coherent images.
    Samples obtained from a model trained on ImageNet@256.
    }
    \label{fig:noise_threshold}
\end{figure*}

\mypar{Noise threshold}
In \Cref{fig:noise_threshold}, we illustrate the impact of using a higher noise threshold (which amounts to using \lpl for longer in the diffusion chain) on the image quality.
A higher noise threshold yields better structures in the images and exacerbates semantic features that distinguish objects.

\begin{figure*}
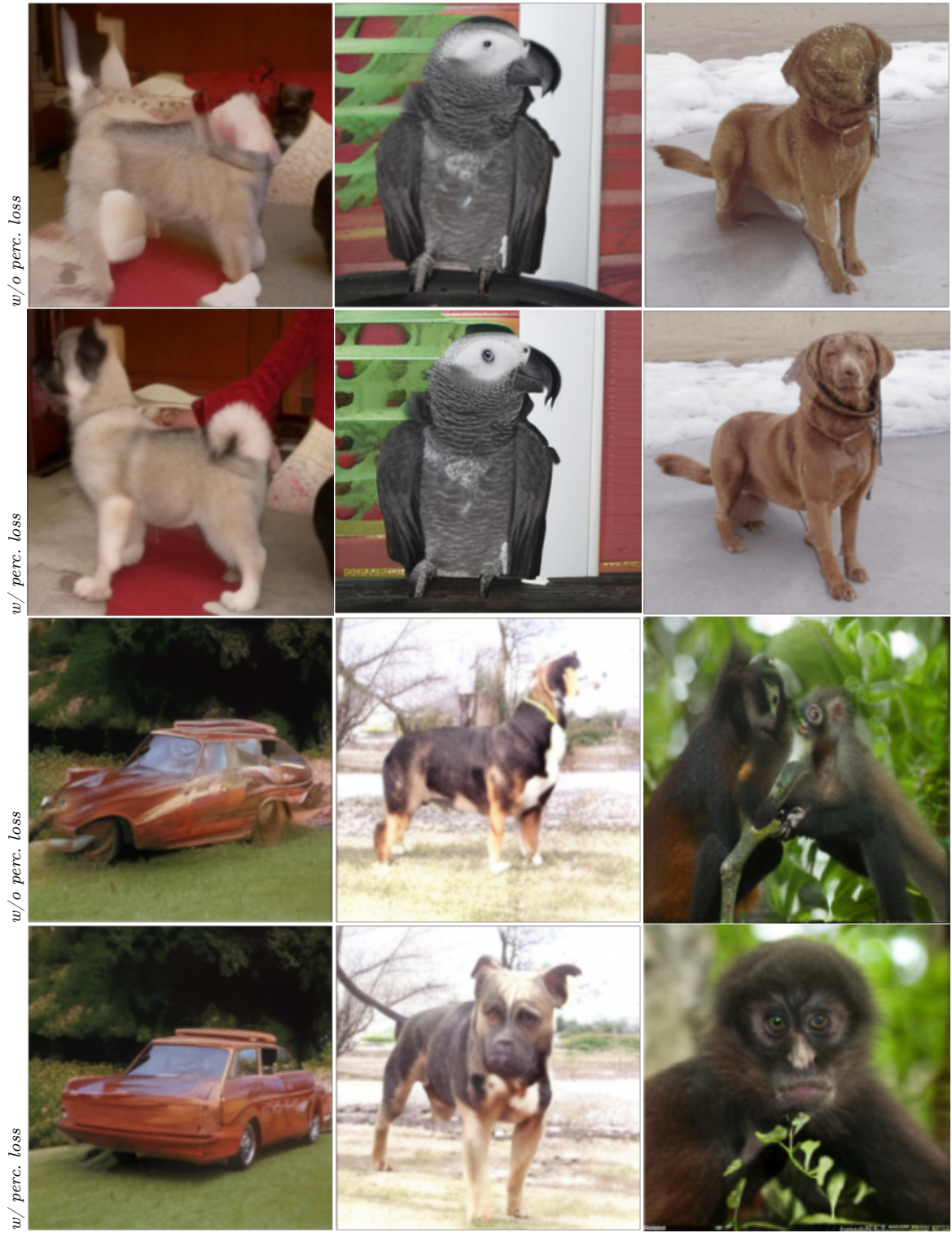

    \def\myim#1#2{\includegraphics[width=50mm,height=50mm]{figures/imagenet_256_base/#1/#2.png}}
    \centering
    \setlength\tabcolsep{0.5pt}
    \renewcommand{\arraystretch}{0.2}
    \begin{tabular}{cccc}
    \begin{sideways} {\it \smaller w/o perc.\ loss} \end{sideways}   
     & \myim{baseline}{baseline-1}     & \myim{baseline}{baseline-2}  
     & \myim{baseline}{baseline-3}
\\
     \begin{sideways} {\it \smaller w/ perc.\ loss} \end{sideways}   
     & \myim{lpl}{lpl-1}     & \myim{lpl}{lpl-2}  
     & \myim{lpl}{lpl-3}
     \\
     \begin{sideways} {\it \smaller w/o perc.\ loss} \end{sideways}   
     & \myim{baseline}{baseline-4}
     & \myim{baseline}{baseline-5}     & \myim{baseline}{baseline-6}
\\
     \begin{sideways} {\it \smaller w/ perc.\ loss} \end{sideways}   
     & \myim{lpl}{lpl-4}
     & \myim{lpl}{lpl-5}     & \myim{lpl}{lpl-6}
     \\
    \end{tabular}
    \caption{{\bf Qualitative comparison of the effect of the latent peceptual loss.} 
    Models trained on \imnet at 256 resolution with (bottom) and without (top) our perceptual loss. 
    Without the perceptual loss, the model frequently fails to generate coherent structures, using the perceptual loss, the model generates more plausible objects with sharper details. The models are finteuned for 100k iterations from a checkpoint that was trained for 200k iterations. The samples are generated {\it without classifier-free  guidance or EMA}, using 50 DDIM steps.
    }
    \label{fig:lpl_comp0}
\end{figure*}

\mypar{Vanilla diffusion}
In \Cref{fig:lpl_comp0}, we  qualitatively investigate the influence of LPL on a baseline model, without classifier-free guidance and without EMA.
We can see that \lpl significantly improves the structure of objects as compared to the model that was trained without it. 

\begin{figure*}
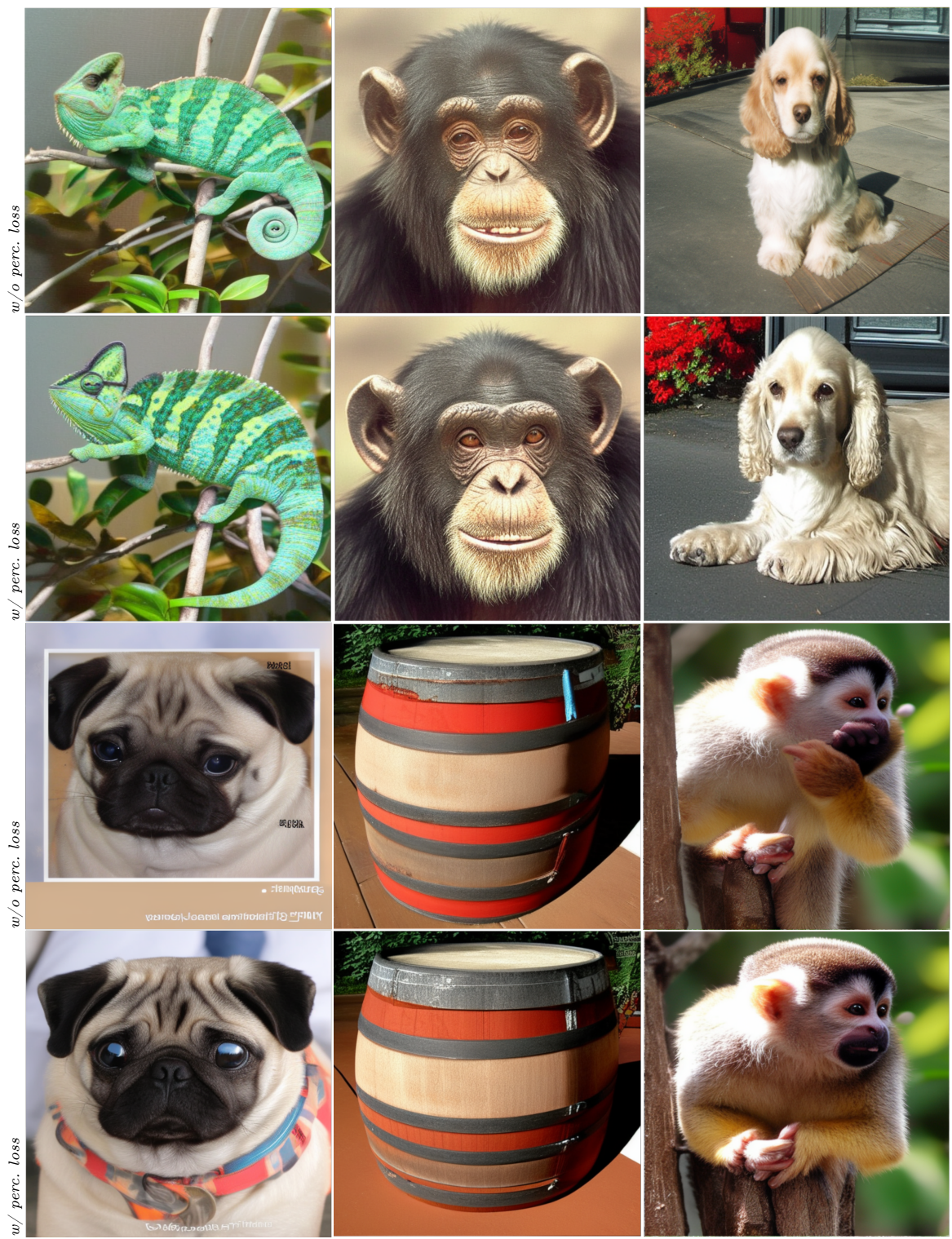

    \def\myim#1#2{\includegraphics[height=50mm,width=50mm]{figures/imagenet_512/#1/#2.png}}
    \footnotesize
    \centering
    \setlength\tabcolsep{0.5pt}
    \renewcommand{\arraystretch}{0.2}
    \begin{tabular}{cccc}
    \begin{sideways} {\it \smaller w/o perc.\ loss} \end{sideways}   
     & \myim{nolpl_image}{nolpl_image-2}  
     & \myim{nolpl_image}{nolpl_image-3}     & \myim{nolpl_image}{nolpl_image-4}
\\
     \begin{sideways} {\it \smaller w/ perc.\ loss} \end{sideways}   
     & \myim{lpl_image}{lpl_image-2}  
     & \myim{lpl_image}{lpl_image-3}     & \myim{lpl_image}{lpl_image-4}
     \\
     \begin{sideways} {\it \smaller w/o perc.\ loss} \end{sideways}
     & \myim{nolpl_image}{nolpl_image-5}     & \myim{nolpl_image}{nolpl_image-6}
     & \myim{nolpl_image}{nolpl_image-7} 
\\
     \begin{sideways} {\it \smaller w/ perc.\ loss} \end{sideways}   
     & \myim{lpl_image}{lpl_image-5}     & \myim{lpl_image}{lpl_image-6}
     & \myim{lpl_image}{lpl_image-7}
     \\
    \end{tabular}
    \caption{{\bf Influence of finetuning a class-conditional model of \imnet at 512 resolution  using our perceptual loss.}
    Our perceptual loss (bottom row) leads to more realistic textures and more detailed images. 
    }
    \label{fig:imagenet_512}
\end{figure*}

\mypar{Samples of \imnet models}
In \Cref{fig:imagenet_512} we show samples of models trained with or without \lpl on \imnet at 512 resolution.
At higher resolutions, we also observe that the model trained with LPL generates images that are sharper and present more fine-grained details compared to the baseline.

\begin{figure*}[t]
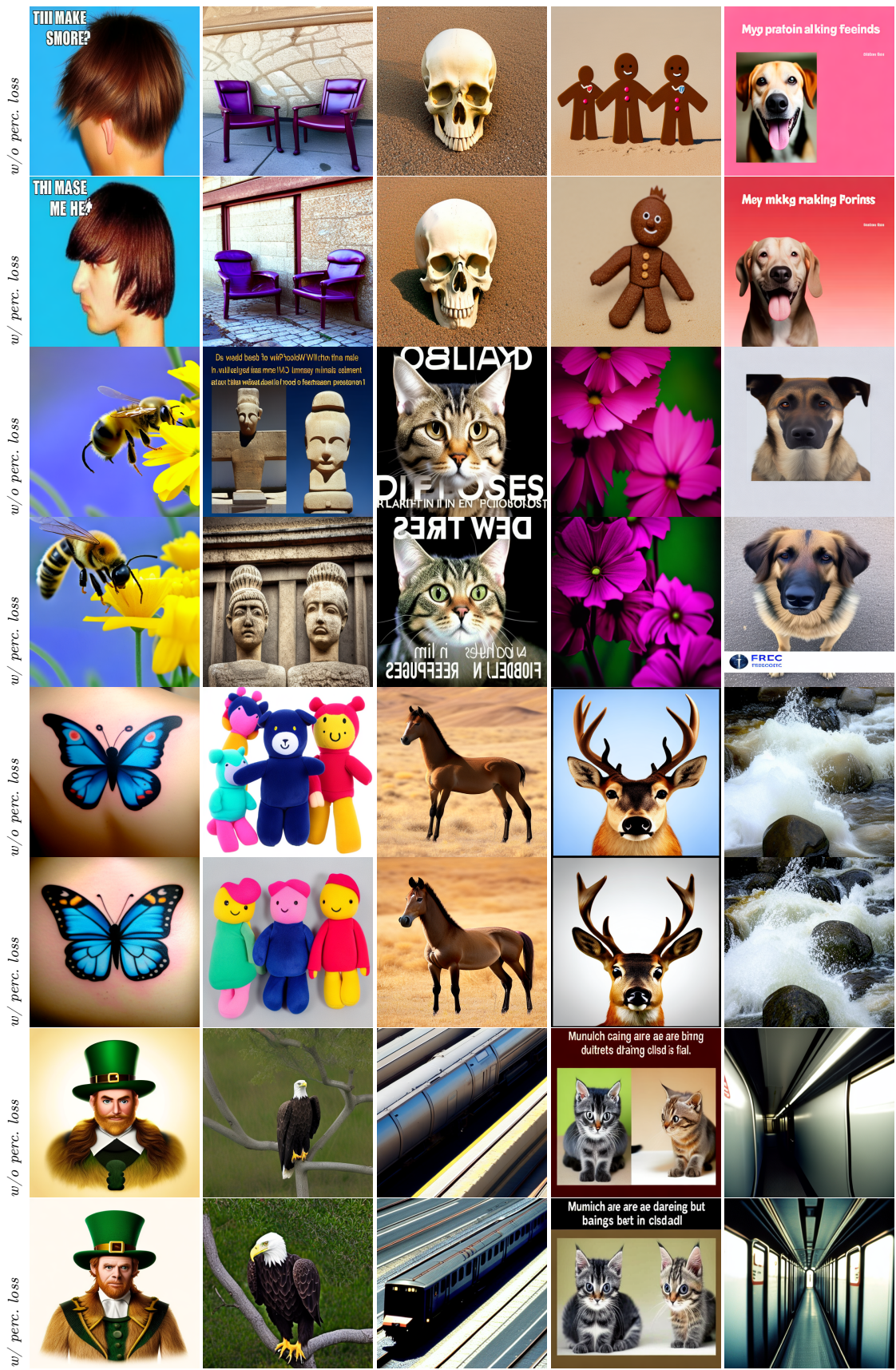

    \def\myim#1#2{\includegraphics[width=31.5mm,height=31.5mm]{figures/CC12M_512/#1/#2.png}}
    \footnotesize
    \centering
    \setlength\tabcolsep{2pt}
    \renewcommand{\arraystretch}{0.2}
     \resizebox{0.9\textwidth}{!}{
    \begin{tabularx}{\linewidth}{cYYYYY}
     \begin{sideways} {\it w/o perc.\ loss} \end{sideways}   
     & \myim{no_lpl}{000006}     & \myim{no_lpl}{000020}  
     & \myim{no_lpl}{000053}     & \myim{no_lpl}{000087}
     & \myim{no_lpl}{000088}     
     \\
     \begin{sideways} {\it w/ perc.\ loss} \end{sideways}   
     & \myim{lpl}{000006}     & \myim{lpl}{000020}  
     & \myim{lpl}{000053}     & \myim{lpl}{000087}
     & \myim{lpl}{000088}     
\\
     \begin{sideways} {\it w/o perc.\ loss} \end{sideways}   
     & \myim{no_lpl}{000120}     & \myim{no_lpl}{000145}  
     & \myim{no_lpl}{000170} 
     & \myim{no_lpl}{000152}     & \myim{no_lpl}{000278}
     \\
     \begin{sideways} {\it w/ perc.\ loss} \end{sideways}   
     & \myim{lpl}{000120}     & \myim{lpl}{000145}  
     & \myim{lpl}{000170}     
     & \myim{lpl}{000152}     & \myim{lpl}{000278}
\\
     \begin{sideways} {\it w/o perc.\ loss} \end{sideways}     
     & \myim{no_lpl}{000357}     & \myim{no_lpl}{000112}
     & \myim{no_lpl}{000497}     & \myim{no_lpl}{000575}
     & \myim{no_lpl}{000060}
     \\
     \begin{sideways} {\it w/ perc.\ loss} \end{sideways} 
     & \myim{lpl}{000357}     & \myim{lpl}{000112}
     & \myim{lpl}{000497}     & \myim{lpl}{000575}
     & \myim{lpl}{000060}
\\
     \begin{sideways} {\it w/o perc.\ loss} \end{sideways}   
     & \myim{no_lpl}{000761}     & \myim{no_lpl}{000783}  
     & \myim{no_lpl}{000994}     & \myim{no_lpl}{000837}
     & \myim{no_lpl}{000669}
     \\
     \begin{sideways} {\it w/ perc.\ loss} \end{sideways}   
     & \myim{lpl}{000761}     & \myim{lpl}{000783}  
     & \myim{lpl}{000994}     & \myim{lpl}{000837}
     & \myim{lpl}{000669}
\\
    \end{tabularx}}
    \caption{{\bf Qualitative comparison.} Comparison of samples from models trained with and without our perceptual loss on CC12M at $512$ resolution (The differences are better viewed by zooming in).}
    \label{fig:lpl_cc12m_ext}
\end{figure*}

\begin{figure*}[t]
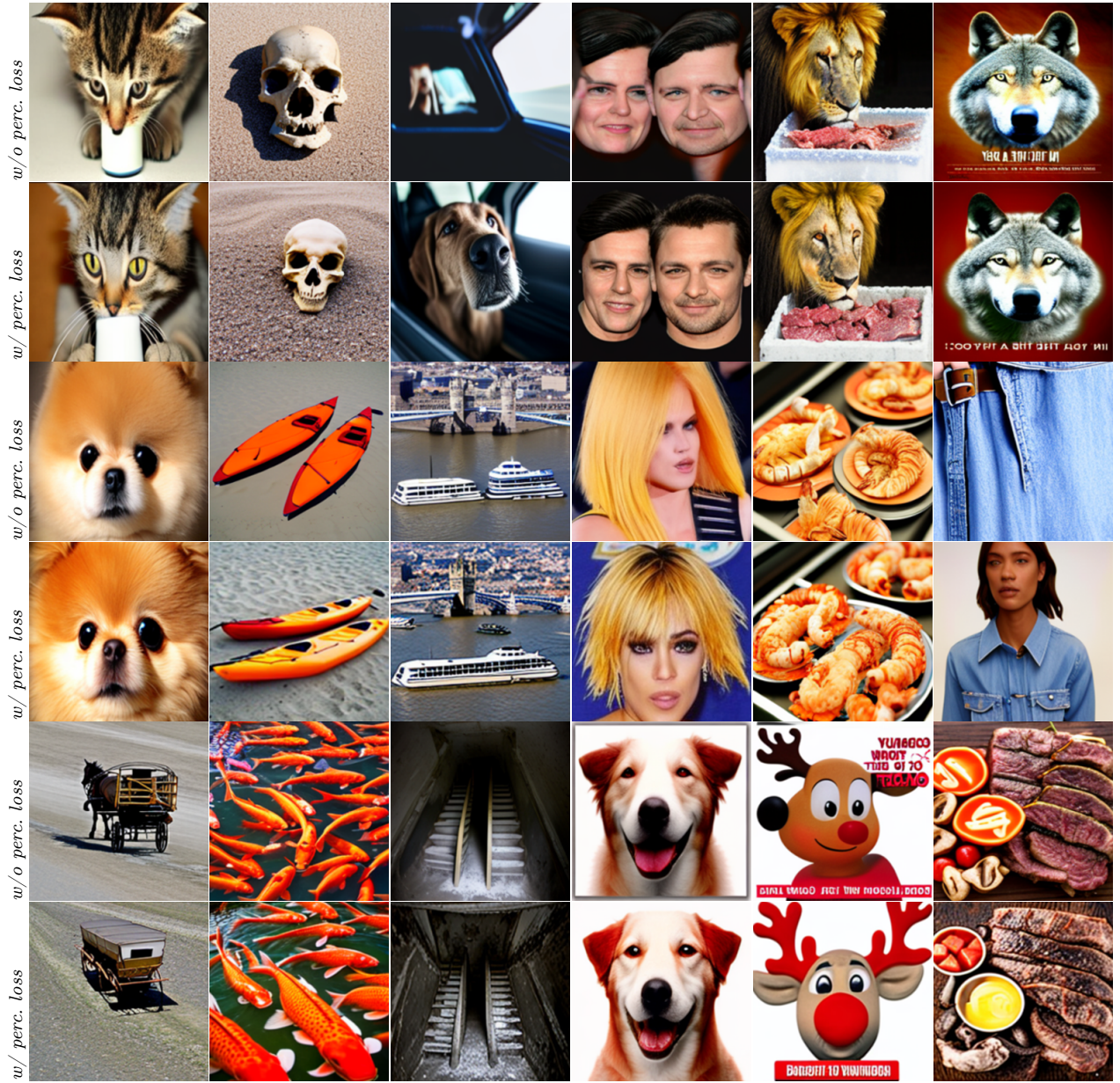

    \def\myim#1#2{\includegraphics[width=27mm,height=27mm]{figures/images_t2i_256/#1/image/image-#2.png}}
    \footnotesize
    \centering
    \setlength\tabcolsep{0.5pt}
    \renewcommand{\arraystretch}{0.2}
    \begin{tabular}{ccccccc}
     \begin{sideways} {\it w/o perc.\ loss} \end{sideways}   
     & \myim{baseline}{1}     & \myim{baseline}{2}  
     & \myim{baseline}{3}     & \myim{baseline}{4}
     & \myim{baseline}{5}     & \myim{baseline}{6}
     \\
     \begin{sideways} {\it w/ perc.\ loss} \end{sideways}   
     & \myim{lpl}{1}     & \myim{lpl}{2}  
     & \myim{lpl}{3}     & \myim{lpl}{4}
     & \myim{lpl}{5}     & \myim{lpl}{6}
\\
     \begin{sideways} {\it w/o perc.\ loss} \end{sideways}   
     & \myim{baseline}{7}     & \myim{baseline}{8}  
     & \myim{baseline}{9}     & \myim{baseline}{10}
     & \myim{baseline}{11}     & \myim{baseline}{12}
     \\
     \begin{sideways} {\it w/ perc.\ loss} \end{sideways}   
     & \myim{lpl}{7}     & \myim{lpl}{8}  
     & \myim{lpl}{9}     & \myim{lpl}{10}
     & \myim{lpl}{11}     & \myim{lpl}{12}
\\
     \begin{sideways} {\it w/o perc.\ loss} \end{sideways}   
     & \myim{baseline}{13}     & \myim{baseline}{14}  
     & \myim{baseline}{15}     & \myim{baseline}{16}
     & \myim{baseline}{17}     & \myim{baseline}{18}
     \\
     \begin{sideways} {\it w/ perc.\ loss} \end{sideways}   
     & \myim{lpl}{13}     & \myim{lpl}{14}  
     & \myim{lpl}{15}     & \myim{lpl}{16}
     & \myim{lpl}{17}     & \myim{lpl}{18}
\\
    \end{tabular}
    \caption{{\bf Qualitative comparison.} of samples from models trained with and without our \lpl on \shst at $256$ resolution. }
    \label{fig:lpl_ss_ext}
\end{figure*}

\mypar{Samples on T2I models}
We provide additional qualitative comparisons regarding our \lpl loss.
\Cref{fig:lpl_cc12m_ext} showcases results on a model trained on \cc at $512$ resolution, \Cref{fig:lpl_ss_ext} showcases results on a model trained on \shst at $256$ resolution.

\mypar{Captions for \Cref{fig:lpl_cc12m_ext}}

\begin{enumerate}
    \item this makes me miss my short hair.
    \item person weathered , cracked purple leather chairs sitting outside of a building.
    \item human skull on the sand.
    \item gingerbread little men on the beach.
    \item my dog , person making friends.
    \item a bee gathering nectar from a wild yellow flower.
    \item two - headed statue in an ancient city of unesco world heritage site.
    \item the - cat breeds in photographs.
    \item a beautiful shot of the flowers.
    \item  photo of rescue dog , posted on the page on facebook.
    \item blue butterfly tattoo on back of the shoulder.
    \item soft toy in a choice of colours.
    \item wild mustang spring foal with its mare in a parched alpine meadow.
    \item surprised buck with wide eyes.
    \item water rushing through rocks in a river.
    \item the legend of the leprechaun.
    \item bald eagle in a tree.
    \item tilt up to show an elevated train riding down the track.
    \item munchkin cats are gaining in popularity , but is breeding these cats cruel?
    \item interiors of a subway train.
\end{enumerate}

\end{document}